\documentclass{article}
\PassOptionsToPackage{numbers, compress}{natbib}
 \usepackage[preprint]{neurips_2026}

\usepackage[utf8]{inputenc} % allow utf-8 input
\usepackage[T1]{fontenc}    % use 8-bit T1 fonts
\usepackage{hyperref}       % hyperlinks
\usepackage{url}            % simple URL typesetting
\usepackage{booktabs}       % professional-quality tables
\usepackage{amsfonts}       % blackboard math symbols
\usepackage{nicefrac}       % compact symbols for 1/2, etc.
\usepackage{microtype}      % microtypography
\usepackage{xcolor}         % colors
\usepackage{graphicx}
\usepackage{xspace}
\usepackage{booktabs}
\usepackage{multirow}
\usepackage{makecell}
\usepackage{xcolor}
\usepackage{colortbl}
\usepackage{adjustbox}
\usepackage{booktabs}
\usepackage{longtable}
\usepackage{xcolor}
\usepackage{colortbl}
\usepackage{makecell}
\usepackage{array}
\usepackage{wrapfig}
\usepackage{amsmath}
\usepackage{subcaption}
\usepackage{tcolorbox}
\usepackage{mdframed}

\newcommand{\huggingfacedown}{\includegraphics[height=0.75em]{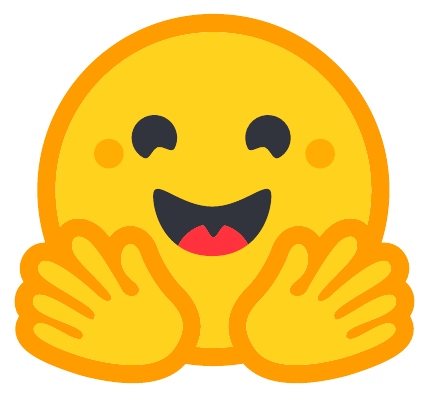}\xspace}
\newcommand{\githubdown}{\includegraphics[height=0.75em]{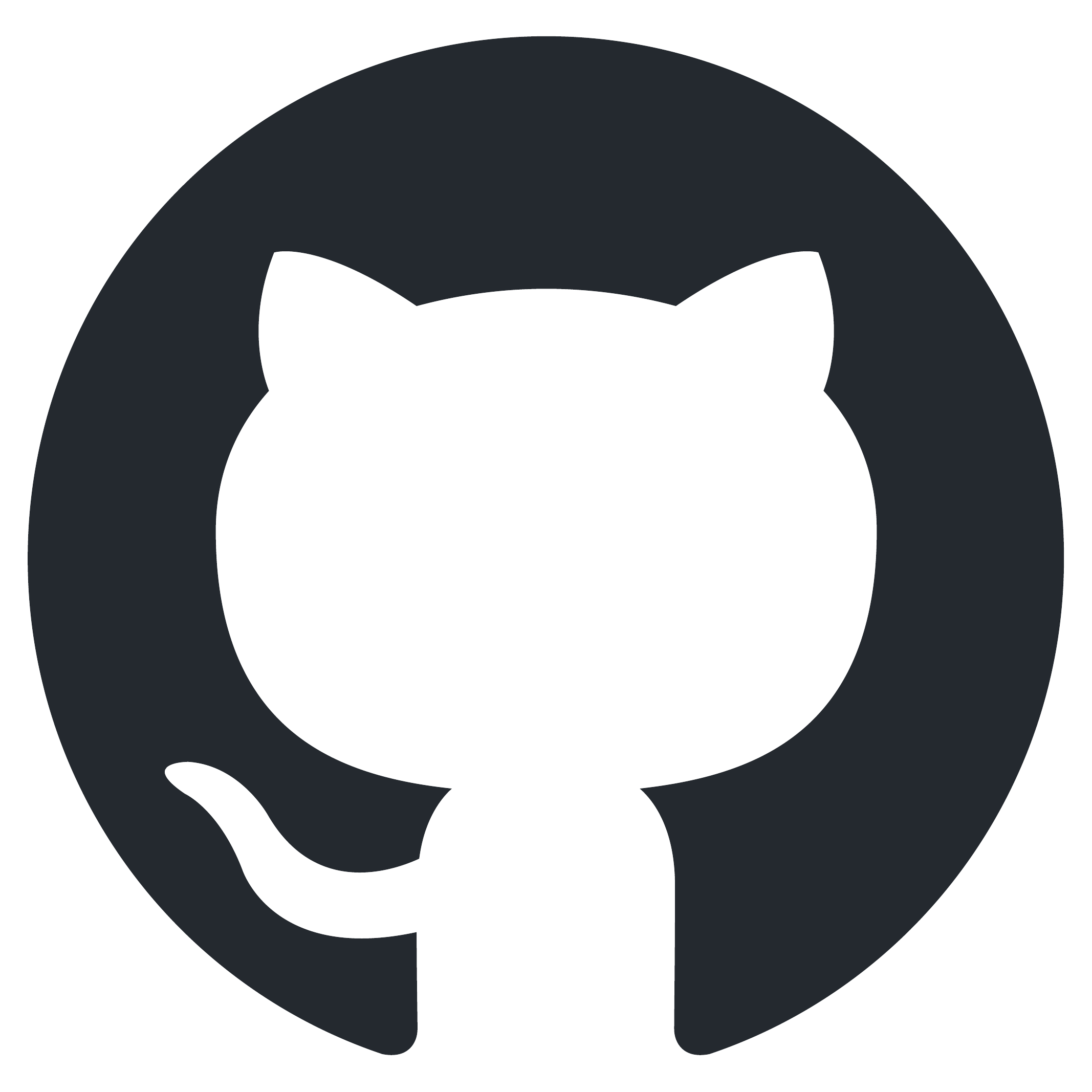}\xspace}
\newcommand{\name}{GlotOCR\allowbreak\ Bench\xspace}

\title{\name: OCR Models Still Struggle Beyond a Handful of Unicode Scripts}

\author{\textbf{Amir Hossein Kargaran}$^{1,3}$\thanks{Correspondence to \texttt{amir@cis.lmu.de}} \quad  \textbf{Nafiseh Nikeghbal}$^{2,3}$ \\ 
\textbf{Jana Diesner}$^{2,3}$ \quad
\textbf{François Yvon}$^{4}$ \quad\textbf{Hinrich Schütze}$^{1,3}$ 
\\\\
$^{1}$LMU Munich \quad 
$^{2}$TU Munich \quad
$^{3}$MCML \\
$^{4}$Sorbonne Université \& CNRS, ISIR \protect
\\\\
\githubdown Pipeline Code: \url{https://github.com/cisnlp/glotocr-bench}\\
\huggingfacedown Benchmark: \url{https://hf.co/datasets/cis-lmu/glotocr-bench}
\vspace{-2em}}

\begin{document}

\maketitle

\begin{abstract}
Optical character recognition (OCR) has advanced rapidly with the rise of vision-language
models, yet evaluation has remained concentrated on a small cluster of high- and mid-resource
scripts. We introduce \name, a comprehensive benchmark evaluating OCR
generalization across 100+ Unicode scripts. Our benchmark comprises clean and degraded
image variants rendered from real multilingual texts.
Images are rendered using fonts from the Google Fonts repository, shaped with HarfBuzz and rasterized with FreeType, supporting both LTR and RTL scripts. Samples of rendered images were manually reviewed to verify correct rendering across all scripts.
We evaluate a broad suite of open-weight and proprietary vision-language models and
find that most perform well on fewer than ten scripts, and even the strongest frontier
models fail to generalize beyond thirty scripts. Performance broadly tracks script-level pretraining coverage, suggesting that current OCR systems rely on language model pretraining as much as on visual recognition.
Models confronted with unfamiliar scripts either produce random noise or
hallucinate characters from similar scripts they already know. We release the benchmark and pipeline for reproducibility.
\end{abstract}

\section{Introduction}
\label{sec:intro}

Optical character recognition (OCR) is among the oldest problems in pattern recognition, yet
the field's evaluation practices have quietly narrowed over time. The dominant
benchmarks, e.g., OCRBench~\citep{liu2024ocrbench}, OCRBench
v2~\citep{fu2025ocrbenchv2}, CC-OCR~\citep{yang2025cc}, and
OmniDocBench~\citep{ouyang2025omnidocbench}, evaluate models on Latin, CJK, and a small number of other mid-resource scripts. Even recent work explicitly targeting multilingual OCR, such as
\citep{li2025dots} with its XDocParse benchmark (not publicly released) covering 126~languages, focuses on
\emph{languages} rather than \emph{scripts}, and the underlying script diversity
remains limited. Work on minority scripts~\citep{liu2026omniocr} has made
important progress, but covers only a handful of writing systems. No existing benchmark
evaluates OCR across the full breadth of Unicode.

This matters because the Unicode Standard (version 17.0 at the time of writing) currently encodes 172 scripts, representing thousands of years of human writing across every inhabited continent.
Many of these
scripts are still in active use by millions of speakers; others are of critical
importance to historical linguistics, archaeology, and cultural preservation~\citep{diao2025ancient}. The digitization of documents in these scripts depends on OCR, yet we have no systematic picture of where current models succeed and where they fail entirely. Moreover, digitized books and scanned documents represent a largely untapped source of training data for low-resource and historical languages --- one that recent initiatives are beginning to exploit~\citep{kydlicek2025finepdfs} but that requires reliable OCR across scripts to be of use.

\begin{wrapfigure}{r}{0.55\linewidth}
  \centering
  \includegraphics[width=\linewidth]{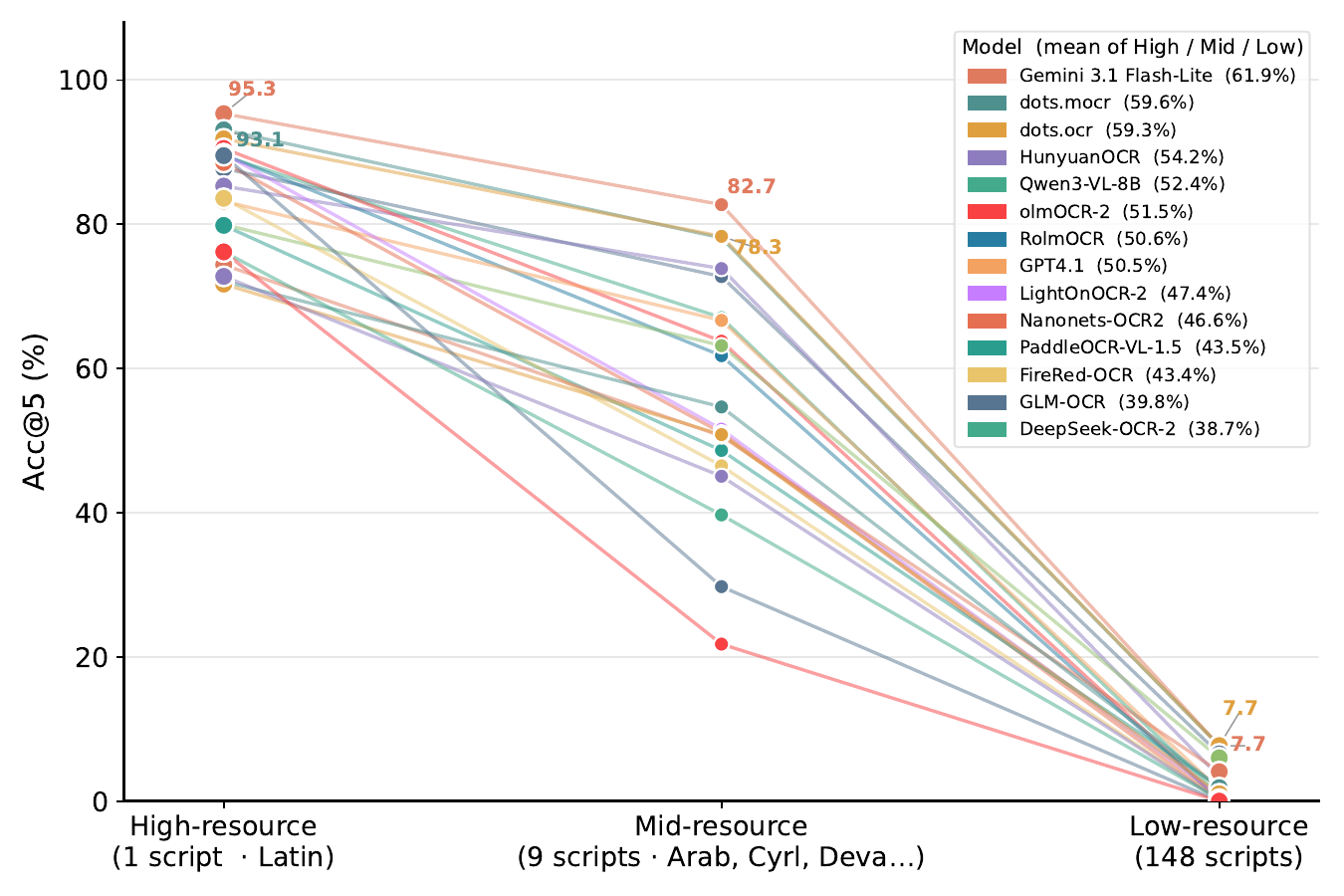}
\caption{Acc@5 (see~\S\ref{sec:metrics}) by script resource tier (high, mid, low). Performance drops sharply on low-resource scripts. Gemini~3.1 Flash-Lite leads (95.3\%, 82.7\%) but falls to 7.7\%; others score <1\%.}
  \label{fig:resource_tiers}
  \vspace{-2pt}
\end{wrapfigure}

We fill this gap with \name, a benchmark spanning 158 Unicode
scripts with clean and degraded image variants, carefully curated text, and rendering. Our evaluation of a broad suite of open-weight and API-based
vision-language models reveals
findings
that are both simple and striking.
Figure~\ref{fig:resource_tiers} summarizes them directly.

Every model evaluated, whether a frontier API system or a specialized open-weight OCR model, performs well on Latin and degrades substantially on mid-resource scripts such as Arabic, Cyrillic, and Devanagari. On the remaining 148 scripts, which constitute 94\% of the scripts in our benchmark, performance falls below 10\% Acc@5 (see~\S\ref{sec:metrics}) for all models; the best result is 7.7\%, achieved by Gemini~3.1 Flash-Lite, dots.ocr and dots.mocr, while most models score below 1\%.
These 148 scripts are not obscure edge cases: they include scripts actively used by tens of millions of people (Ethiopic, Khmer, Sinhala), scripts central to entire national literatures (Armenian, Tibetan, Myanmar), and scripts whose digitization is essential for linguistic and cultural preservation (Linear~B, N'Ko, Vai, and dozens of historical writing systems)~\citep{daniels1996world}. Beyond poor transcription accuracy, model failure on low-resource scripts is not "silent": rather than refusing or indicating uncertainty, models tend to produce fluent-looking text in a script they do know.
A model confronted with Gujarati may output Devanagari; confronted with Thaana, it may produce Arabic. 
This pattern suggests that pretraining coverage plays a central role alongside
visual recognition: models recognize that an image contains text, but map it
onto the nearest script in their training data.
We make the following contributions:
\begin{itemize}
  \item {\name}, a \textbf{benchmark} covering 158 Unicode scripts with clean
        and degraded image variants, sampled from real multilingual texts, rendered with script-aware font selection and proper
        bidirectional handling.

    \item A \textbf{comprehensive evaluation} of open-weight and frontier OCR models, reporting character error rate and acceptance rate (Acc@0 and Acc@5) stratified by script and resource level.

    \item \textbf{Core findings}: (a)~OCR generalization is effectively restricted to a handful of scripts for all current models; (b)~performance broadly tracks script-level pretraining coverage, with a steep performance drop between mid- and low-resource scripts; and (c)~models confronted with unfamiliar scripts hallucinate characters from known scripts rather than failing silently.

  \item A \textbf{public release} of the benchmark dataset, rendering pipeline, evaluation code, and per-model results.

\end{itemize}

%% ============================================================
\section{Related Work}
\label{sec:related}

\paragraph{OCR benchmarks.}
OCR benchmarks have historically concentrated on a small set of high- and mid-resource
scripts. Recent multimodal benchmarks have expanded the range of evaluated tasks
but remain narrow in script coverage. OCRBench~\citep{liu2024ocrbench} and its
successor OCRBench v2~\citep{fu2025ocrbenchv2} evaluate large multimodal models
across document parsing, key information extraction, and multilingual
recognition, but coverage remains largely limited to Latin and Chinese scripts.
CC-OCR~\citep{yang2025cc} covers eleven languages including Arabic, Japanese,
Korean, and Vietnamese. Reasoning-OCR~\citep{he2025reasoning} probes logical reasoning from OCR
cues. OmniDocBench~\citep{ouyang2025omnidocbench} benchmarks end-to-end
document parsing in English and Chinese. olmOCR-Bench~\citep{poznanski2025olmocr}
focuses on English-language PDF parsing. KITAB-Bench~\citep{heakl-etal-2025-kitab}
targets Arabic OCR and document understanding specifically.
OCRTurk~\citep{yilmaz2026ocrturk} addresses Turkish specifically.
\citet{sohail2024deciphering} benchmark LLM-based OCR for low-resource scripts
including Urdu and Tajik, finding that performance degrades with text length.
\citet{dasanaike2026sococrbench} introduce socOCRbench, a private benchmark
targeting social science documents across multiple world regions and scripts;
the script coverage is broader than most benchmarks  and their model rankings are broadly consistent with ours, but the dataset is not
publicly available.
These last two works are the closest in spirit to ours, yet none of the benchmarks
discussed above treats script coverage as the primary axis of evaluation.

\paragraph{OCR datasets and low-resource adaptation.}
\citet{agarwal-anastasopoulos-2024-concise} survey OCR techniques for
low-resource languages with a focus on indigenous languages of the Americas,
identifying data scarcity and script support as key open challenges.
For Indic scripts, \citet{saini2022ocr} introduce a synthetic dataset across
23 Indic languages, and \citet{kolavi-etal-2025-nayana} propose a LoRA-based
adaptation framework for ten Indic languages using synthetic data.
\citet{sarkar2024printed} address printed OCR for extremely low-resource Indic languages, introducing synthetic and real word-level datasets for nine Indian languages. CAMIO~\citep{arrigo-etal-2022-camio} covers 35 languages across 24
scripts as a data resource with
transcriptions for only 13 languages, and is available only through the LDC
catalog at cost. OmniOCR~\citep{liu2026omniocr} introduces dynamic LoRA
adaptation for minority scripts, applied to only four writing systems:
Tibetan, Shui, Ancient Yi, and Dongba, though limited to single-character
classification
datasets~\citep{yuan2018tibetanmnist,liu2024ancient,luo2023multiple}.
dots.ocr~\citep{li2025dots} and its XDocParse benchmark span 126 languages,
yet script diversity is not the primary axis. Historical document OCR has
received attention from \citet{greif2025multimodal}, but only for Latin-script
historical documents. These efforts demonstrate growing awareness of the
problem but address individual script families rather than Unicode coverage as a
whole. \name is the first benchmark to evaluate OCR generalization across
most of the Unicode script inventory.

\section{\name}
\label{sec:dataset}

\name covers 158 Unicode scripts with two image variants per sentence:
a \emph{clean} rendering on a white background and a \emph{degraded} rendering
simulating aged documents. Sentences are sampled from real multilingual text
and rendered into images, which are then presented to OCR systems for
transcription.
Rendering text into images is a common practice in OCR dataset
construction~\citep{yim2021synthtiger,malik2026synthocr}; \citet{malik2026synthocr}
for instance, create synthetic OCR data for Kashmiri at the word level. 
Our benchmark extends
this approach to a far broader set of scripts, using sentence-level text where
available and falling back to word-level text for scripts with limited data.

We sample up to 100 sentences per script, with the exception of Latin, for which we sample 4,000 sentences, and a small set of mid-resource scripts, for which we sample 400 sentences to enable per-language analysis within a single script. The total number of sentences is 16,375. For 68 scripts we have fewer than 100 sentences (and words); for these, virtually all models fail even at the script identification level (ScriptAcc), as shown in Table~\ref{tab:zero_sa}. While the small sample sizes for these scripts limit the conclusions that can be drawn, the consistent failure across virtually all models suggests that this reflects genuine model limitations on low-resource scripts rather than evaluation artifact. We cap evaluation at 100 sentences per script to keep evaluation cost low while maximizing script coverage.
The resource tier assigned to each script is based on their prevalence in web content.\footnote{\url{https://en.wikipedia.org/wiki/Languages\_used\_on\_the_Internet}.}
Three tiers are defined: High (Latin script only), Mid (nine scripts with substantial pretraining coverage: Arabic, Cyrillic, Devanagari, Han, Japanese, Hangul, Greek, Hebrew, and Thai), and Low (the remaining 148 scripts).

\subsection{Text}

Text data was assembled from multiple sources to achieve the broadest possible Unicode
script coverage. The
primary text collection is the GlotLID v3
dataset collection~\citep{kargaran-etal-2023-glotlid,kargaran2024glotcc}, which covers over
2,102 language--script pairs from different sources. We prioritize sentences between 30 and 100 characters  (not so short as to be trivial, not so long as to overwhelm the model) from publicly available sources whose licenses do not restrict public sharing.

For scripts with insufficient GlotLID coverage, we gathered additional data from: Wiktionary~\citep{wiktionary}; WikiSource~\citep{wikisource2026}; Omniglot~\citep{ager2026omniglot} sample texts; Google Fonts language data~\citep{googlefonts}; and texts converted into different writing systems for scripts with limited native digital text~\citep{aksharamukha_web, karamolegkou-etal-2025-evaluating}. We also draw on Common Crawl–derived datasets: the \texttt{und\_*} labels in GlotLID model denote scripts for which no genuine multilingual data exists, but whose script can still be predicted---for example, \texttt{und\_Sylo} refers to text in the Sylheti script. We retrieve data for such labels from GlotCC~\citep{kargaran2024glotcc} and FineWeb2~\citep{penedo2025fineweb}, which applied this model to Common Crawl data. We filter entries from other scripts and map them to their primary language where possible (e.g., \citet{sefat2026glotweb} map \texttt{und\_Sylo} to \texttt{syl\_Sylo}, since only one language is conventionally written in that script).

For all newly gathered sources, the Unicode script of each sentence was verified using GlotScript~\citep{kargaran-etal-2024-glotscript} to ensure that the languages represented are conventionally written in those scripts. We did not consider randomly generated character sequences for rare scripts; while such synthetic text may be useful for training data augmentation, it lacks the linguistic validity required for evaluation. % The additional data gathered for this benchmark will be incorporated into the forthcoming GlotLID v4.0 release.

\subsection{Image}\label{sec:font_render}

\setlength{\textfloatsep}{5pt}
\begin{wraptable}{r}{0.49\columnwidth}\vspace{-5mm}
  \centering
  \caption{{Font availability per resource tier. Counts are per font family. \vspace{-2pt}}}
  \footnotesize
  \label{tab:font_stats}
  \resizebox{0.48\columnwidth}{!}{%
  \begin{tabular}{l r r r r r}
  \toprule
  Tier & Scripts & Total  & Median & Min & Max \\
  \midrule
    High & 1 & 1907  & 1907 & 1907 & 1907 \\
    Mid & 9 & 814  & 59 & 29 & 323 \\
    Low & 148 & 380 & 1 & 1 & 29 \\
  \midrule
    \textbf{All} & 158 & 3101 & 1 & 1 & 1907 \\
  \bottomrule
  \end{tabular}}
\end{wraptable}
\textbf{Font.}
All fonts were sourced from the Google Fonts Files~\citep{googlefonts} under
SIL Open Font License v1.1.\footnote{\url{https://github.com/google/fonts/tree/main/ofl}}
We wrote a script to sort all fonts by the Unicode scripts they cover, using the
metadata provided by the repository.
Font metadata alone is insufficient to guarantee correct rendering: a font may
declare support for a script yet fail to render specific codepoints. For each
sentence, the rendering font is selected through the following pipeline:
(1)~filter to fonts declared for the script of that sentence;
(2)~among those, retain only fonts that cover all Unicode codepoints in the
sentence;
(3)~then retain only fonts that successfully render all glyphs in the sentence;
(4)~select one randomly from the remaining candidates.
All three filtering conditions are necessary: our manual audit revealed cases
where a font declared codepoint support but failed at the rendering stage (solved by step 3).
Finally, we manually inspected ten rendered images per script across a range of image sizes to confirm visual correctness. For common scripts, this was
verified against external editors; for rare scripts where no editor renders them reliably, we verified character by character against the Unicode
character charts.

\textbf{Rendering.}
Images are rendered using HarfBuzz~\citep{harfbuzz2026} for text shaping and
FreeType~\citep{freetype2024} for glyph rasterization. Sentences with mixed
bidirectional content are excluded; all rendered text is uniformly LTR or RTL.
For each sentence we produce two image variants. The \emph{clean} variant renders
text on a plain white canvas with slight random rotation. The \emph{degraded}
variant applies a pipeline of augmentations simulating an aged physical document:
textured paper backgrounds, ink spread and wear, geometric distortions, resolution
downsampling, and JPEG compression artifacts. These operations reflect common
artifacts in document capture and scanning pipelines~\citep{groleau2022augraphy}
and standard augmentation practices in OCR
literature~\citep{gupta2016synthetic,yim2021synthtiger}.
Full rendering parameters are given in Appendix~\ref{app:rendering}.
Appendix Figure~\ref{fig:sample_images} displays representative examples for Greek and Aghwan
scripts. Note that vertically written scripts such as Mongolian are treated as
horizontal text, as vertical rendering is not supported by our pipeline.

\textbf{Release.}
The dataset is publicly released on Hugging Face under an evaluation-only license: the benchmark may be used for testing models
but may not be used in any form, or derivative thereof, for training. The
rendering pipeline is released separately under the Apache 2.0 license and may
be used to generate training data from different seed text. The release includes
clean and degraded image variants, ground-truth transcriptions, and metadata (script, language, and text source).

\section{Evaluation Setup}
\label{sec:eval}\label{sec:metrics}
\textbf{Evaluation pipeline.} All models are evaluated in zero-shot mode using the \texttt{uv-scripts/ocr}
inference suite~\citep{uvscripts2026ocr}. Where a chat template is available
it is applied; otherwise the prompt is passed directly. The prompt simply asks
the model to transcribe the text in the image, return it wrapped in tags, and
provide no commentary or explanation. Images are provided at their native
rendered resolution without further preprocessing.

\textbf{Models.} We evaluate the following open-weight OCR models:
dots.ocr~\citep{li2025dots},
dots.mocr (dots.ocr-1.5)~\citep{zheng2026multimodalocr},
olmOCR-2~\citep{poznanski2025olmocr2},
RolmOCR~\citep{rolmocr2025},
LightOnOCR-2~\citep{taghadouini2026lightonocr},
Nanonets-OCR2~\citep{Nanonets-OCR2},
PaddleOCR-VL-1.5~\citep{cui2026paddleocr},
FireRed-OCR~\citep{wu2026firered},
GLM-OCR~\citep{duan2026glm},
DeepSeek-OCR-2~\citep{wei2026deepseekocr2},
HunyuanOCR~\citep{hunyuan2025hunyuanocr}, and
Qwen3-VL-8B~\citep{bai2025qwen3vl}.
We additionally evaluate two proprietary models via their respective APIs: Gemini~3.1 Flash-Lite~\citep{gemini31_flash_lite_blog_2026,comanici2025gemini}
and GPT-4.1~\citep{openai2024gpt4}. As the field evolves rapidly, we maintain
an online leaderboard where we plan to add additional models, including Falcon OCR~\citep{bevli2026falconperception},
Qianfan-OCR~\citep{dong2026qianfan},
MonkeyOCR~\citep{li2025monkeyocr},
Nemotron OCR v2~\citep{nvidia_nemotron_ocr_v2},
and
Chandra OCR 2~\citep{chandra_ocr_2_2026}.

\textbf{Metrics.}
We use three metrics throughout the paper. Our primary metric is CER, defined as the normalized Levenshtein edit distance at the character level with whitespace ignored: $\text{CER} = \min\!\left(1, \frac{S + D + I}{N}\right)$, where $S$, $D$, and $I$ are respectively the substitution, deletion, and insertion counts, and $N$ is the ground-truth length. To account for minor expected variations in model output, we report the best CER across four configurations: the original output, the reversed output string, the lowercased output, and the version with Unicode combining marks removed. We additionally report Acc@$k$ (A@$k$ for short): the fraction of sentences for which
$\text{CER} \leq k/100$, with $k \in \{0, 5\}$. Acc@5 is our primary
accuracy metric, measuring near-perfect transcription and mapping naturally
onto the binary question of whether a model can operate in a given script.
Finally, Script Accuracy (ScriptAcc) measures whether the model responds in
the correct script regardless of transcription accuracy, as determined by
GlotScript~\citep{kargaran-etal-2024-glotscript}.
Throughout this paper, scripts
are identified by their ISO~15924 four-letter codes (e.g., \texttt{Arab} for
Arabic).
This metric disentangles
script identification from transcription quality and serves as a diagnostic
for cross-script hallucination.

\section{Results}
\label{sec:results}

\begin{table}[t]
\centering
\caption{
\name benchmark results by resource tier.
Each tier result is the macro average over its scripts.
A@0 and A@5 denote the fraction of predictions with CER $\leq 0$ and $\leq 0.05$, respectively. Bold = best; underline = 2nd best}
\label{tab:main_results}
\vspace{4pt}
\footnotesize
\resizebox{0.90\textwidth}{!}{%
\begin{tabular}{l | ccc | ccc | ccc | ccc}
\toprule
\multirow{2}{*}{\textbf{Model}}
  & \multicolumn{3}{c|}{\textbf{High} (1 script)}
  & \multicolumn{3}{c|}{\textbf{Mid} (9 scripts)}
  & \multicolumn{3}{c|}{\textbf{Low} (148 scripts)}
  & \multicolumn{3}{c} {\textbf{Mean (across tiers)}}
\\
  & \makecell{CER\\$\downarrow$} & \makecell{A@0\\$\uparrow$} & \makecell{A@5\\$\uparrow$} & \makecell{CER\\$\downarrow$} & \makecell{A@0\\$\uparrow$} & \makecell{A@5\\$\uparrow$} & \makecell{CER\\$\downarrow$} & \makecell{A@0\\$\uparrow$} & \makecell{A@5\\$\uparrow$} & \makecell{CER\\$\downarrow$} & \makecell{A@0\\$\uparrow$} & \makecell{A@5\\$\uparrow$} \\
\midrule
  Gemini 3.1 Flash-Lite & \cellcolor[HTML]{FFFFFF}$\mathbf{0.9}$ & \cellcolor[HTML]{85C1E9}$\mathbf{86.0}$ & \cellcolor[HTML]{85C1E9}$\mathbf{95.3}$ & \cellcolor[HTML]{FFFFFF}$\mathbf{3.0}$ & \cellcolor[HTML]{85C1E9}$\mathbf{66.1}$ & \cellcolor[HTML]{85C1E9}$\mathbf{82.7}$ & \cellcolor[HTML]{FFFFFF}$\mathbf{79.0}$ & \cellcolor[HTML]{8AC4EA}$5.0$ & \cellcolor[HTML]{85C1E9}$\mathbf{7.7}$ & \cellcolor[HTML]{FFFFFF}$\mathbf{27.7}$ & \cellcolor[HTML]{85C1E9}$\mathbf{52.4}$ & \cellcolor[HTML]{85C1E9}$\mathbf{61.9}$ \\
  dots.mocr & \cellcolor[HTML]{FDF4F5}$\underline{1.5}$ & \cellcolor[HTML]{91C7EB}$\underline{82.5}$ & \cellcolor[HTML]{93C8EB}$\underline{93.1}$ & \cellcolor[HTML]{FDF4F5}$6.0$ & \cellcolor[HTML]{9BCCED}$\underline{57.0}$ & \cellcolor[HTML]{8FC6EA}$78.1$ & \cellcolor[HTML]{FAD9DC}$84.1$ & \cellcolor[HTML]{88C2E9}$\underline{5.1}$ & \cellcolor[HTML]{85C1E9}$\mathbf{7.7}$ & \cellcolor[HTML]{FCEBEC}$30.5$ & \cellcolor[HTML]{97CAEC}$\underline{48.2}$ & \cellcolor[HTML]{91C7EB}$\underline{59.6}$ \\
  dots.ocr & \cellcolor[HTML]{FDF0F1}$1.6$ & \cellcolor[HTML]{98CAEC}$80.4$ & \cellcolor[HTML]{9BCCED}$91.8$ & \cellcolor[HTML]{FEF8F8}$\underline{5.0}$ & \cellcolor[HTML]{9FCEED}$55.4$ & \cellcolor[HTML]{8FC6EA}$\underline{78.3}$ & \cellcolor[HTML]{FBE4E7}$\underline{82.6}$ & \cellcolor[HTML]{85C1E9}$\mathbf{5.2}$ & \cellcolor[HTML]{85C1E9}$\mathbf{7.7}$ & \cellcolor[HTML]{FDF0F1}$\underline{29.7}$ & \cellcolor[HTML]{9CCCED}$47.0$ & \cellcolor[HTML]{92C8EB}$59.3$ \\
  HunyuanOCR & \cellcolor[HTML]{F9CCD1}$3.4$ & \cellcolor[HTML]{EAF4FB}$56.0$ & \cellcolor[HTML]{C5E1F4}$85.3$ & \cellcolor[HTML]{FDF3F4}$6.3$ & \cellcolor[HTML]{A7D2EF}$52.5$ & \cellcolor[HTML]{99CBEC}$73.9$ & \cellcolor[HTML]{F7C2C7}$87.3$ & \cellcolor[HTML]{D6EAF7}$1.7$ & \cellcolor[HTML]{C9E3F5}$\underline{3.4}$ & \cellcolor[HTML]{FBDEE1}$32.3$ & \cellcolor[HTML]{C8E3F5}$36.8$ & \cellcolor[HTML]{ADD5F0}$54.2$ \\
  Qwen3-VL-8B & \cellcolor[HTML]{FCE8EA}$2.0$ & \cellcolor[HTML]{B1D7F1}$72.8$ & \cellcolor[HTML]{AAD4EF}$89.5$ & \cellcolor[HTML]{FDF0F1}$7.2$ & \cellcolor[HTML]{B3D8F1}$47.6$ & \cellcolor[HTML]{A9D3EF}$67.1$ & \cellcolor[HTML]{F5AFB6}$89.8$ & \cellcolor[HTML]{F8FBFD}$0.3$ & \cellcolor[HTML]{F2F8FC}$0.8$ & \cellcolor[HTML]{FAD9DD}$33.0$ & \cellcolor[HTML]{B9DBF2}$40.3$ & \cellcolor[HTML]{B6DAF2}$52.4$ \\
  olmOCR-2 & \cellcolor[HTML]{FCE8EA}$2.0$ & \cellcolor[HTML]{AAD4EF}$75.0$ & \cellcolor[HTML]{A3D0EE}$90.5$ & \cellcolor[HTML]{FCEDEE}$8.3$ & \cellcolor[HTML]{B9DBF2}$45.3$ & \cellcolor[HTML]{B0D7F0}$63.8$ & \cellcolor[HTML]{F5ADB4}$90.2$ & \cellcolor[HTML]{FBFDFE}$0.2$ & \cellcolor[HTML]{FAFCFE}$0.3$ & \cellcolor[HTML]{FAD6DA}$33.5$ & \cellcolor[HTML]{B9DBF2}$40.2$ & \cellcolor[HTML]{BBDCF2}$51.5$ \\
  RolmOCR & \cellcolor[HTML]{FCE8EA}$2.0$ & \cellcolor[HTML]{B2D8F1}$72.7$ & \cellcolor[HTML]{A9D3EF}$89.6$ & \cellcolor[HTML]{FCE7E9}$10.0$ & \cellcolor[HTML]{BBDCF2}$44.6$ & \cellcolor[HTML]{B5D9F1}$61.8$ & \cellcolor[HTML]{F4A0A8}$92.0$ & \cellcolor[HTML]{FBFDFE}$0.1$ & \cellcolor[HTML]{FBFCFE}$0.2$ & \cellcolor[HTML]{F9CED2}$34.7$ & \cellcolor[HTML]{BEDEF3}$39.1$ & \cellcolor[HTML]{C0DFF3}$50.6$ \\
  GPT4.1 & \cellcolor[HTML]{FADADD}$2.7$ & \cellcolor[HTML]{E1F0F9}$58.7$ & \cellcolor[HTML]{D2E8F6}$83.2$ & \cellcolor[HTML]{FDF3F4}$6.3$ & \cellcolor[HTML]{B8DBF2}$45.6$ & \cellcolor[HTML]{AAD3EF}$66.7$ & \cellcolor[HTML]{F9CCD0}$85.9$ & \cellcolor[HTML]{F0F7FC}$0.6$ & \cellcolor[HTML]{E5F2FA}$1.6$ & \cellcolor[HTML]{FBE3E5}$31.7$ & \cellcolor[HTML]{D0E7F6}$35.0$ & \cellcolor[HTML]{C1DFF3}$50.5$ \\
  LightOnOCR-2 & \cellcolor[HTML]{FCE5E7}$2.2$ & \cellcolor[HTML]{A8D3EF}$75.6$ & \cellcolor[HTML]{A8D2EF}$89.8$ & \cellcolor[HTML]{FBDDE0}$13.0$ & \cellcolor[HTML]{E4F1FA}$28.4$ & \cellcolor[HTML]{CCE5F5}$51.6$ & \cellcolor[HTML]{F4A2AA}$91.6$ & \cellcolor[HTML]{FAFCFE}$0.2$ & \cellcolor[HTML]{F3F9FC}$0.7$ & \cellcolor[HTML]{F8C8CD}$35.6$ & \cellcolor[HTML]{D1E7F6}$34.7$ & \cellcolor[HTML]{D1E7F6}$47.4$ \\
  Nanonets-OCR2 & \cellcolor[HTML]{FBE2E4}$2.3$ & \cellcolor[HTML]{B9DBF2}$70.7$ & \cellcolor[HTML]{B0D6F0}$88.6$ & \cellcolor[HTML]{FBDFE2}$12.2$ & \cellcolor[HTML]{D5EAF7}$34.1$ & \cellcolor[HTML]{CDE6F6}$51.1$ & \cellcolor[HTML]{F4A0A8}$91.9$ & \cellcolor[HTML]{FBFDFE}$0.1$ & \cellcolor[HTML]{FCFDFE}$0.2$ & \cellcolor[HTML]{F8C8CD}$35.5$ & \cellcolor[HTML]{D0E7F6}$35.0$ & \cellcolor[HTML]{D5E9F7}$46.6$ \\
  PaddleOCR-VL-1.5 & \cellcolor[HTML]{F6B2B9}$4.6$ & \cellcolor[HTML]{E7F2FA}$57.0$ & \cellcolor[HTML]{E7F3FA}$79.8$ & \cellcolor[HTML]{F5AEB5}$26.9$ & \cellcolor[HTML]{D6EAF7}$33.9$ & \cellcolor[HTML]{D3E8F7}$48.6$ & \cellcolor[HTML]{F4A1A9}$91.8$ & \cellcolor[HTML]{DDEDF8}$1.5$ & \cellcolor[HTML]{DFEFF9}$2.0$ & \cellcolor[HTML]{F4A2AA}$41.1$ & \cellcolor[HTML]{E2F0F9}$30.8$ & \cellcolor[HTML]{E5F2FA}$43.5$ \\
  FireRed-OCR & \cellcolor[HTML]{F9CCD0}$3.4$ & \cellcolor[HTML]{E0EFF9}$59.2$ & \cellcolor[HTML]{CFE7F6}$83.6$ & \cellcolor[HTML]{F8C7CC}$19.3$ & \cellcolor[HTML]{E5F2FA}$27.9$ & \cellcolor[HTML]{D8EBF8}$46.5$ & \cellcolor[HTML]{F4A0A8}$91.9$ & \cellcolor[HTML]{FDFEFE}$0.1$ & \cellcolor[HTML]{FCFDFE}$0.2$ & \cellcolor[HTML]{F6B6BC}$38.2$ & \cellcolor[HTML]{E9F4FB}$29.0$ & \cellcolor[HTML]{E6F2FA}$43.4$ \\
  GLM-OCR & \cellcolor[HTML]{FCE7E9}$2.1$ & \cellcolor[HTML]{B8DBF2}$70.9$ & \cellcolor[HTML]{AAD3EF}$89.5$ & \cellcolor[HTML]{F49FA8}$31.1$ & \cellcolor[HTML]{FFFFFF}$17.8$ & \cellcolor[HTML]{FFFFFF}$29.8$ & \cellcolor[HTML]{F4A5AD}$91.2$ & \cellcolor[HTML]{FFFFFF}$0.0$ & \cellcolor[HTML]{FFFFFF}$0.0$ & \cellcolor[HTML]{F49FA8}$41.5$ & \cellcolor[HTML]{E7F3FA}$29.5$ & \cellcolor[HTML]{F9FCFE}$39.8$ \\
  DeepSeek-OCR-2 & \cellcolor[HTML]{F49FA8}$5.5$ & \cellcolor[HTML]{FFFFFF}$50.1$ & \cellcolor[HTML]{FFFFFF}$76.2$ & \cellcolor[HTML]{F6B6BC}$24.4$ & \cellcolor[HTML]{F3F9FC}$22.2$ & \cellcolor[HTML]{E8F3FA}$39.7$ & \cellcolor[HTML]{F49FA8}$92.0$ & \cellcolor[HTML]{FCFDFE}$0.1$ & \cellcolor[HTML]{FAFCFE}$0.3$ & \cellcolor[HTML]{F4A5AD}$40.6$ & \cellcolor[HTML]{FFFFFF}$24.1$ & \cellcolor[HTML]{FFFFFF}$38.7$ \\
\bottomrule
\end{tabular}
}
\vspace{5pt}
\end{table}

\subsection{Benchmarking OCR Across Resource Tiers}

Table~\ref{tab:main_results} presents results for fourteen OCR systems across
three resource tiers: High (Latin), Mid (9 scripts), and Low (148 scripts),
along with the overall mean. 
On Latin, every model achieves Acc@5 above 75\%
and the top models exceed 90\%.
Performance on mid-resource scripts is
substantially lower but still usable for the best models (Gemini~3.1
Flash-Lite: 82.7\%, dots.ocr: 78.3\%).
The collapse on low-resource scripts
is near-universal: the three best-performing models reach only 7.7\%, representing failure on 92\% of low-resource
sentences.
Overall, Gemini~3.1 Flash-Lite ranks first with CER 27.7 and Acc@5 61.9\%, followed by dots.mocr and dots.ocr. The ranking differences between
models are driven primarily by mid- and low-resource script performance, where
most models fall short.

\textbf{High-resource tier.}
All models perform strongest on the High (Latin script) tier, consistent with
the dominance of Latin-script data in both pretraining corpora and model
development attention. However, no model
achieves near-perfect performance: most models retain a CER above
2\% on Latin, meaning at least 2 errors every 100 characters. One contributing factor is orthographic variation across Latin-script languages;
rare characters such as Icelandic \textit{þ} are frequently confused with
visually similar Latin letters such as \textit{p}~\citep{jasonarson-etal-2023-generating}. 
Such confusions are more prevalent in models trained predominantly on English, which may not have seen sufficient examples of non-English Latin characters during training.

\textbf{Mid-resource tier.} Performance degrades substantially in the Mid tier. Acc@5 drops by 27.6
percentage points on average from High (87.6\%) to Mid (60.0\%), reflecting
both the lower resource level of these scripts and the comparatively less
attention they receive in model development. Gemini~3.1 Flash-Lite leads again
with dots.ocr ranking second. A clear gap emerges below the top systems:
Qwen3-VL-8B and olmOCR-2 fall approximately 15--19 points behind, while
GLM-OCR and DeepSeek-OCR-2 perform over 40 points below Gemini~3.1
Flash-Lite, indicating particularly poor generalization to mid-resource
scripts.

\textbf{Low-resource tier.} Performance degrades severely across
all models in the low tier, reflecting
the extreme scarcity of training data for the remaining 148 scripts. On average,
Acc@5 drops by 57.7 percentage points from the mid tier (60.0\%) to the low tier
(2.3\%), a substantially steeper decline than that observed between the High and Mid tiers. {Gemini~3.1 Flash-Lite}, {dots.ocr} and {dots.mocr} again lead; however,
even these top models achieve Acc@5 scores below 8\%, underscoring that low-resource
script recognition remains largely unsolved. For the remaining models, 11 of the 14 score below 5\% Acc@5 and 8 fall below 1\%, indicating an almost complete failure to generalize to low-resource scripts.

\begin{mdframed}[linecolor=black!70, topline=false, bottomline=false, rightline=false, backgroundcolor=gray!0, innertopmargin=3pt, innerbottommargin=3pt]
\textbf{Finding 1.} OCR performance is largely solved for high-resource Latin script but degrades severely in mid- and low-resource tiers, where even the best models fail on over 92\% of low-resource sentences. The transition from mid- to low-resource is not a smooth degradation but a sharp discontinuity, suggesting a threshold phenomenon: models either have sufficient training exposure to a script or they do not. The gap between top proprietary and best open-weight models is small --- thanks to dots.ocr and dots.mocr --- but most open-weight approaches still lag substantially.
\end{mdframed}

\begin{figure}[t]
  \centering
  \includegraphics[width=0.93\linewidth]{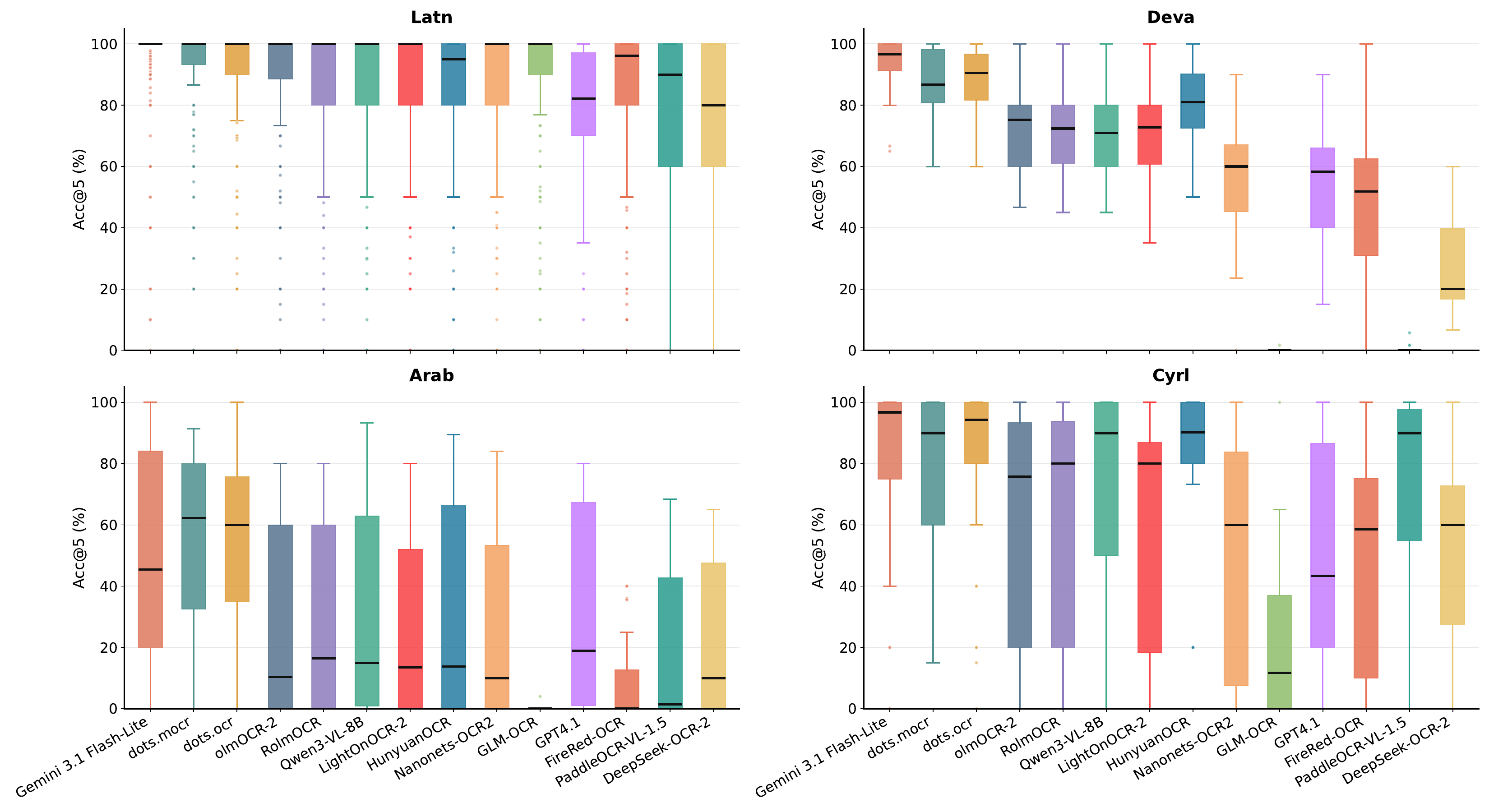}
    \caption{Acc@5 distributions for four scripts (Latin, Devanagari, Arabic, Cyrillic). Boxes correspond to models; each point is the score for one language within the script.}
  \label{fig:latin_boxplot}
  \vspace{3pt}
\end{figure}

\subsection{Per-Language Variance Within Scripts}

Figure~\ref{fig:latin_boxplot} presents the per-language Acc@5 distribution across models for four scripts --- Latin, Devanagari, Arabic, and Cyrillic --- selected as the four scripts with the most languages in the benchmark. For Latin-script languages, models generally achieve high median performance (typically above 90\%), but with notable variability across languages and low-scoring outliers, revealing that strong aggregate performance does not imply uniform coverage across all Latin-script languages.

Performance on non-Latin scripts is generally more variable and degraded. For Devanagari, median accuracies are lower than Latin with moderate spread, though most models maintain moderate performance. Remaining errors are often attributable to conjunct 
characters
--- where multiple
characters
merge into a single glyph --- though models generally handle them well as they are a core feature of the script. Arabic shows the most severe degradation: most models exhibit low medians, wide interquartile ranges, and strong downward skew --- reflecting Arabic's orthographic complexity, where visually similar characters, optional diacritics, and numerous script variations make generalization across its many languages particularly challenging. Cyrillic presents a comparatively stronger pattern, with several models achieving medians comparable to Latin, though variability remains substantial and some models perform poorly.
Across all scripts, models such as {Gemini~3.1 Flash-Lite}, {dots.ocr}, and {dots.mocr} demonstrate tighter distributions and higher medians, indicating more stable cross-language performance --- yet even these models exhibit failures on specific languages.
\begin{mdframed}[linecolor=black!70, topline=false, bottomline=false, rightline=false, backgroundcolor=gray!0, innertopmargin=3pt, innerbottommargin=3pt]
\textbf{Finding 2.} Per-language performance varies substantially within and across scripts. Among the four scripts evaluated, Arabic exhibits the steepest degradation, reflecting its orthographic complexity and the diversity of languages it encodes.
\end{mdframed}

\subsection{Script Accuracy vs.\ OCR Accuracy}

\begin{figure}[t]
  \centering
  \begin{subfigure}[t]{0.47\linewidth}
    \centering
    \includegraphics[width=\linewidth]{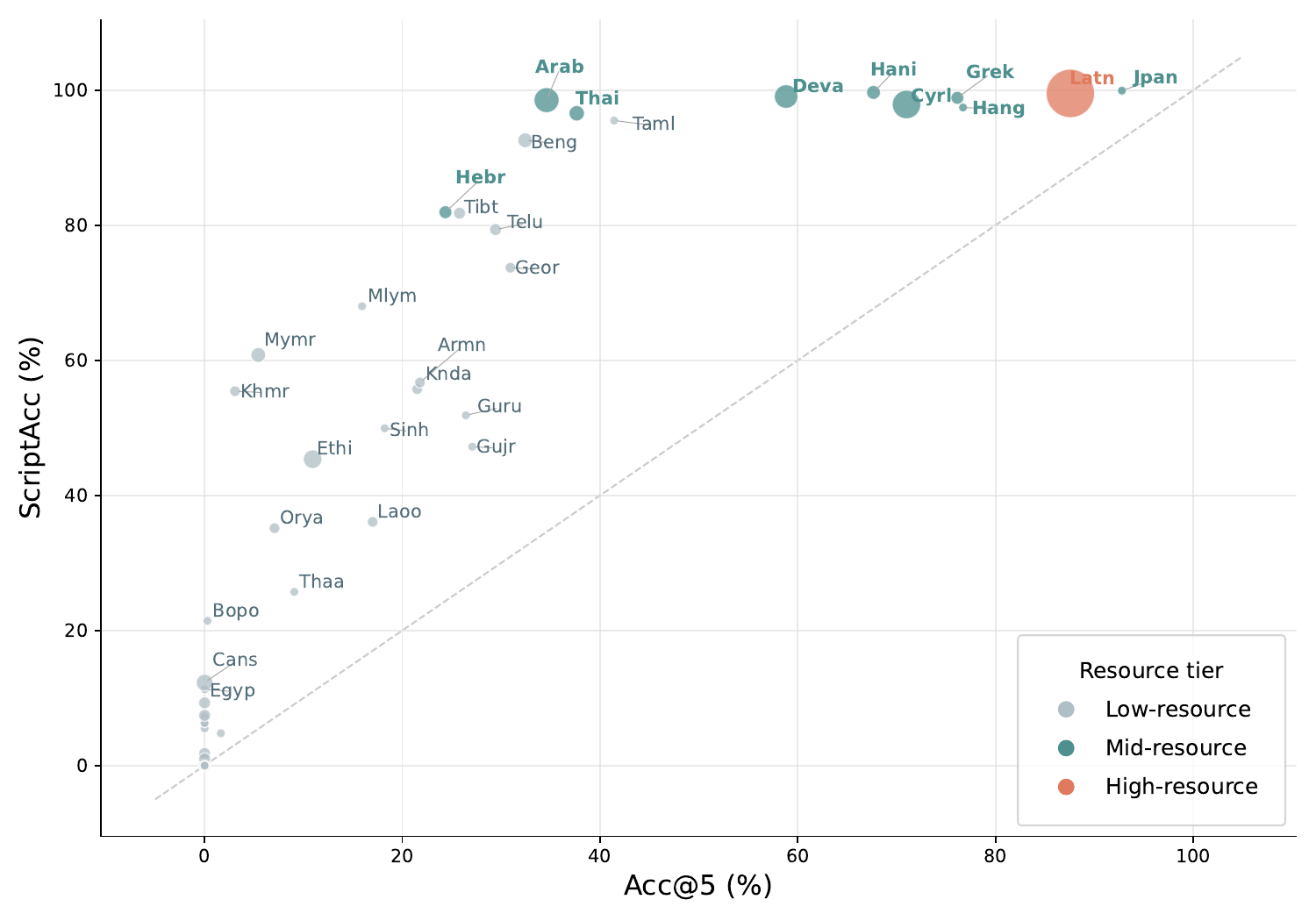}
    \caption{Script-level recognition accuracy (ScriptAcc) vs.\ OCR accuracy
      (Acc@5), averaged across all models. Bubble size $\propto$ log number of
      languages using the script. Resource tier is indicated by color.}
    \label{fig:scriptacc_vs_acc5}
  \end{subfigure}
  \hspace{0.02\linewidth}
  \begin{subfigure}[t]{0.44\linewidth}
    \centering
    \includegraphics[width=\linewidth]{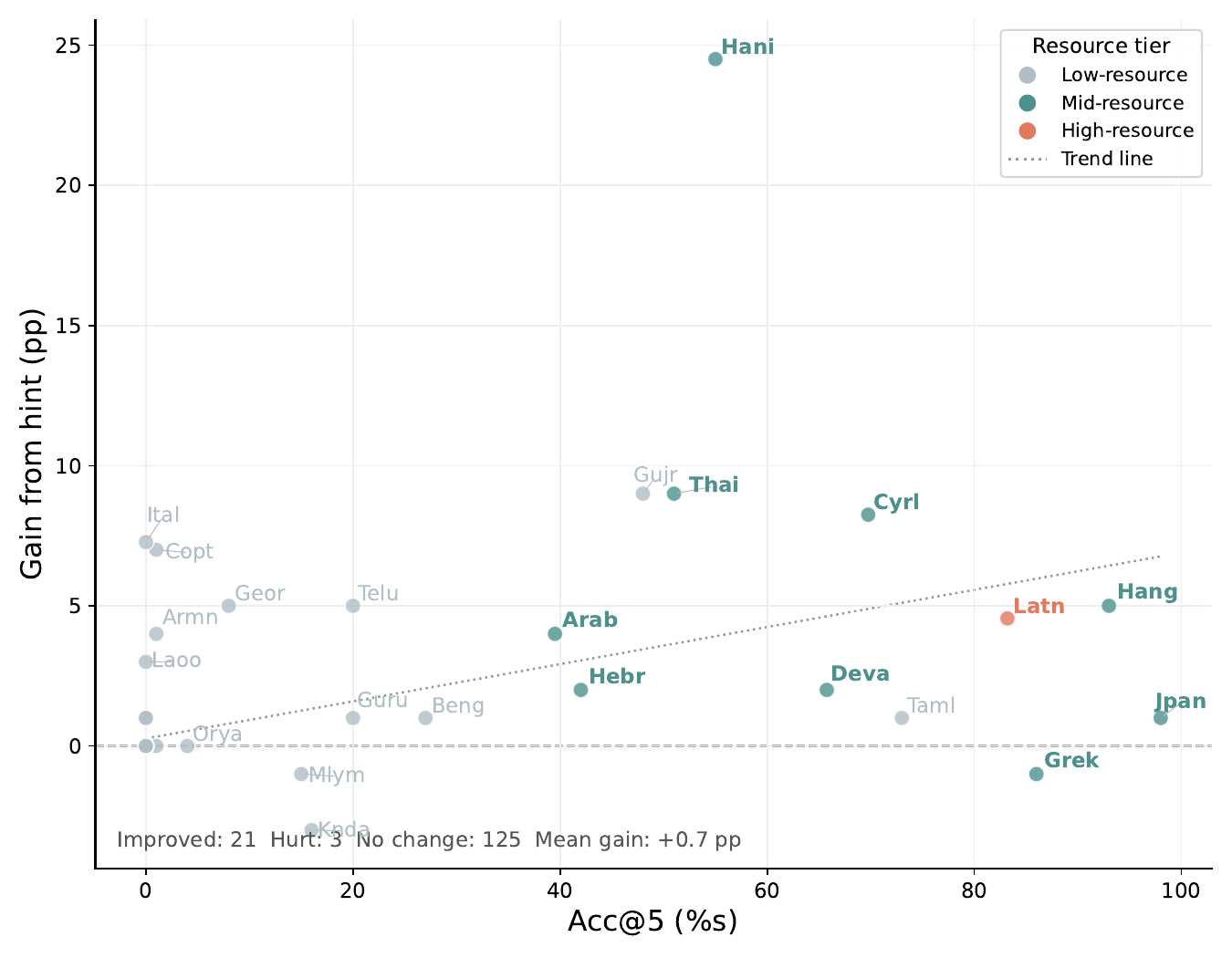}
    \caption{Gain in Acc@5 from script-identity hint vs.\ baseline Acc@5,
      per script for GPT4.1. Most scripts show modest gains
      (+0.7\,pp on average), with Han (Hani) benefiting most.}
    \label{fig:hint_gain_scatter}
  \end{subfigure}
   \caption{Script Recognition and Hint-Guided OCR Analysis.}
  \label{fig:combined}
\end{figure}

Figure~\ref{fig:scriptacc_vs_acc5} presents the relationship between script recognition accuracy ({ScriptAcc}) and OCR accuracy ({Acc@5}), averaged across models. ScriptAcc serves as a prerequisite for Acc@5: models that fail to produce correct script characters cannot achieve high OCR accuracy. This is reflected in the strong diagonal correlation, where resource tier largely determines placement --- high- and mid-resource scripts (Latin, Japanese, Greek, Han) cluster in the upper-right, low-resource scripts occupy the middle band and bottom-left.
Notable deviations reveal additional factors at play. Arabic achieves high ScriptAcc yet lags in Acc@5, suggesting that script-level confusions are not the bottleneck; rather, intra-script variation and visually similar characters drive OCR errors. Hebrew presents a different failure mode: its ScriptAcc is comparatively low due to frequent confusion with Thai (see Table~\ref{tab:script_a}), pulling its OCR performance below scripts like Tamil that suffer less from such cross-script confusion.
Japanese is a notable positive outlier, achieving higher Acc@5 than even Latin despite combining three writing systems --- Hiragana, Katakana, and Kanji. This suggests that OCR models can handle mixed-script sentences well, though we do not investigate code-switching further in this paper. 
\begin{mdframed}[linecolor=black!70, topline=false, bottomline=false, rightline=false, backgroundcolor=gray!0, innertopmargin=3pt, innerbottommargin=3pt]
\textbf{Finding 3.} ScriptAcc is a weak but early indicator of Acc@5. Deviations reveal distinct failure modes: Arabic suffers from intra-script variation despite high ScriptAcc, Hebrew is hurt by cross-script confusion with Thai, and Japanese exceeds Latin in OCR accuracy despite combining three writing systems.
\end{mdframed}

\subsection{Effect of Script-Aware Hinting on OCR Performance}\label{sec:hint}
Figure~\ref{fig:hint_gain_scatter} illustrates the relationship between baseline OCR accuracy (Acc@5) and the gain obtained from providing GPT-4.1 with an explicit hint --- informing the model of the language, script, and the exact characters present in the image, deduplicated and sorted by Unicode code point. Since this provides an unfair advantage for short texts, we exclude samples with fewer than 10 characters, leaving 149 scripts. Out of all evaluated scripts, 125 show no change, 21 improve, and only 3 are negatively affected, yielding a mean gain of +0.7 percentage points --- indicating that script-identity hinting provides selective but limited benefits overall.
Several mid-resource scripts show pronounced improvements: Hani exhibits a gain exceeding 20 percentage points, despite GPT-4.1 already having good ScriptAcc for it. The model tends to produce common tokens rather than visually similar rare characters; providing the exact character set corrects these substitutions. This is expected given Han's large character inventory --- constraining the candidate set has a greater impact when thousands of characters are possible. Cyrillic and Thai also benefit notably, suggesting that character ambiguity is a non-negligible factor for these scripts.
Some low-resource scripts (e.g., Ital, Copt) show modest gains from near-zero baselines, improving to between 5--10\%, which is encouraging but still far from usable performance. Overall, hinting is not the primary bottleneck for most scripts --- particularly in low-resource tier where the model lacks both the underlying visual recognition capability and sufficient pretraining exposure to those characters entirely.
\begin{mdframed}[linecolor=black!70, topline=false, bottomline=false, rightline=false, backgroundcolor=gray!0, innertopmargin=3pt, innerbottommargin=3pt]
\textbf{Finding 4.} Script-aware hinting yields only marginal overall gains (+0.7 pp mean), with 125 of 149 scripts showing no change. Benefits are selective: Hani, Cyrillic, and Thai improve notably due to character ambiguity, while low-resource scripts remain largely unsolved --- indicating that the bottleneck is insufficient visual recognition and pretraining exposure, not script identity.
\end{mdframed}

\subsection{Robustness to Image Degradation Across Resource Tiers}
\begin{figure}[t]
  \centering
  \includegraphics[width=0.99\linewidth]{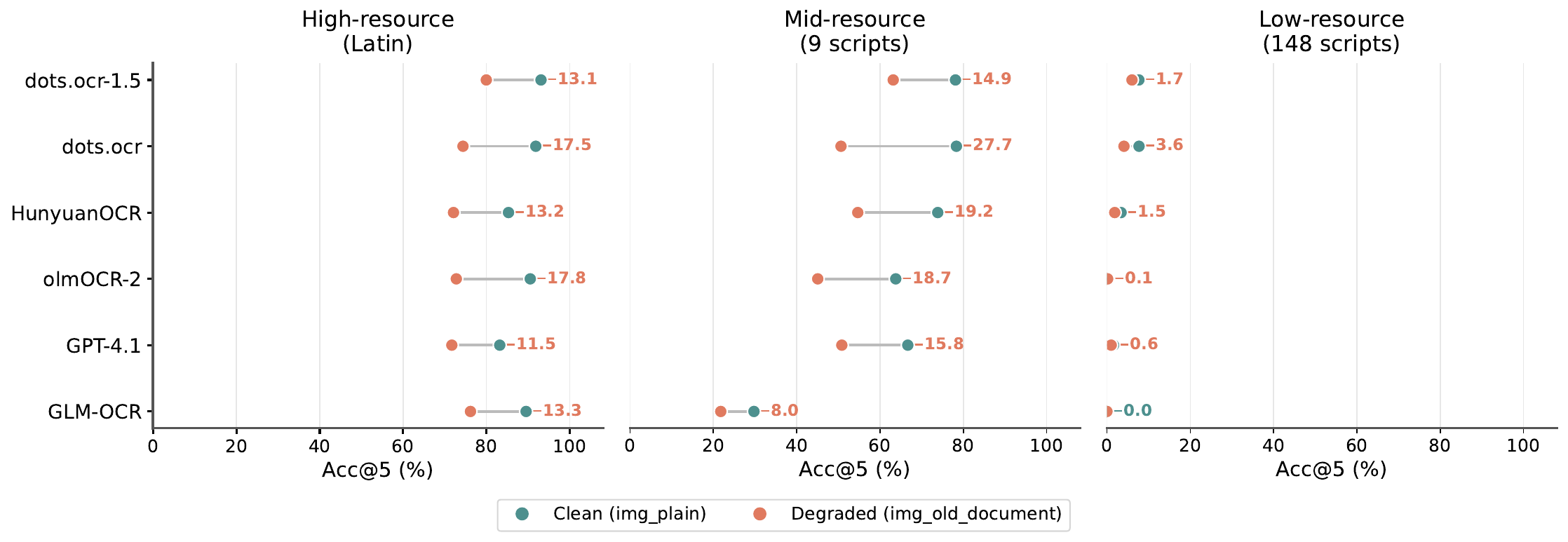}
  \caption{Acc@5 on clean vs. degraded images for the six best-performing models across high-, mid-, and low-resource tiers, with corresponding performance drops.}
  \label{fig:degradation_tiers}
\end{figure}

We evaluate the robustness of the six best-performing models from the previous experiments by comparing their performance on clean images versus degraded ``old document'' variants across three resource tiers. Figure~\ref{fig:degradation_tiers} reports Acc@5 for both conditions, along with the performance drop. Across all models and tiers, performance consistently drops under image degradation, though the magnitude of the loss varies substantially by resource level. In the high-resource tier, all models exhibit noticeable performance drops, indicating that even well-represented scripts remain sensitive to image degradation. GPT-4.1 displays relatively better robustness (13.8\% relative drop), while olmOCR-2 shows a larger decline (19.7\% relative drop).
The mid-resource tier reveals greater sensitivity than high-resource scripts --- lower clean performance and larger absolute drops --- suggesting that models struggle more to recover correct text for lower-resource scripts. This likely reflects reduced familiarity with script-specific visual patterns and weaker generalization under degraded conditions. For low-resource scripts, although absolute degradation is small, relative degradation is greater, given that baseline performance is already near zero.
\begin{mdframed}[linecolor=black!70, topline=false, bottomline=false, rightline=false, backgroundcolor=gray!0, innertopmargin=3pt, innerbottommargin=3pt]
\textbf{Finding 5.} All models suffer performance loss under image degradation across all resource tiers. Clean images represent an upper bound on OCR performance. Latin scripts show consistent but moderate drops, with GPT-4.1 being the most robust among the evaluated models. Sensitivity increases as resource levels decrease.
\end{mdframed}

\subsection{Cross-Script Hallucination}

For each prediction we detect the dominant script of the output and compare it to
the expected script.
We distinguish three failure modes: \emph{cross-script hallucination}, where the model
produces text in a {different} recognizable script; \emph{silence}, where
the model returns an empty or whitespace-only response; and \emph{artifact},
where the output contains characters that GlotScript cannot assign to any real
script---typically repetitive digit strings, punctuation loops, or
model-specific wrapper tokens left over from structured-output formats.

Table~\ref{tab:hallucination_rates} reports the hallucination, silence, and
artifact rates for each model, computed as a macro-average over scripts.
Across all models, only 12.5\% of predictions are
on average assigned to the correct script.
Cross-script hallucination accounts for 68.4\% on average,
artifacts for 13.1\%, and silence for only 6.0\%, showing that models
overwhelmingly prefer to confabulate in a wrong script rather than abstain.
The artifact rate is highly model-dependent:
{DeepSeek-OCR2} (26.2\%) produce far more
artifacts than {dots.ocr} (3.8\%), consistent with models trained to always
emit output---even from blank images. {dots.ocr}, by contrast, mostly chooses
to remain silent when it cannot recognise a script (42.1\%).

\setlength{\textfloatsep}{5pt}
\begin{wraptable}{r}{0.52\columnwidth}\vspace{-5mm}
  \centering
  \caption{Script-level error rates (\%) per model, macro-averaged over scripts, ranked by cross-script hallucination (Hall.) rate.
  The four categories are mutually exclusive and sum to 100\%.}
  \footnotesize
  \label{tab:hallucination_rates}
  \resizebox{0.51\columnwidth}{!}{%
  \begin{tabular}{lcccc}
  \toprule
  \textbf{Model} & \textbf{Correct}~$\uparrow$ & \textbf{Hall.}~$\downarrow$ & \textbf{Silent} & \textbf{Artifact} \\
  \midrule
  {dots.ocr}          & 15.8 & {38.3} & 42.1 & \phantom{0}3.8 \\
  {dots-mocr}         & {16.1} & {50.2} & 16.3 & 17.4 \\
  {FireRed-OCR}       & \phantom{0}8.3 & 62.4 & 12.7 & 16.6 \\
  {DeepSeek-OCR2}     & \phantom{0}6.5 & 63.7 & \phantom{0}3.6 & 26.2 \\
  {Hunyuan-OCR}       & 11.5 & 68.2 & \phantom{0}0.0 & 20.3 \\
  {PaddleOCR-VL-1.5}  & \phantom{0}7.6 & 69.7 & \phantom{0}0.0 & 22.7 \\
  {Qwen3-VL-8B}       & 11.8 & 70.1 & \phantom{0}3.6 & 14.5 \\
  {Gemini-Flash-Lite} & {22.6} & 70.2 & \phantom{0}0.5 & \phantom{0}6.7 \\
  {GPT-4.1}           & 17.6 & 72.1 & \phantom{0}0.9 & \phantom{0}9.4 \\
  {GLM-OCR}           & 10.8 & 74.2 & \phantom{0}3.0 & 12.0 \\
  {Nanonets-OCR2}     & 11.1 & 78.6 & \phantom{0}0.0 & 10.3 \\
  {RolmOCR}           & 13.3 & 78.9 & \phantom{0}0.0 & \phantom{0}7.8 \\
  {LightOn-OCR-2}     & \phantom{0}8.3 & 79.2 & \phantom{0}0.9 & 11.6 \\
  {olmOCR-2}          & 14.4 & 81.7 & \phantom{0}0.0 & \phantom{0}3.9 \\
  \midrule
  {Average}    & {12.5} & {68.4} & {\phantom{0}6.0} & {13.1} \\
  \bottomrule
  \end{tabular}}
\end{wraptable}

Appendix Tables~\ref{tab:zero_sa}--\ref{tab:script_b} report, for each target
script, the two scripts most frequently observed in model outputs. These show
that hallucinated outputs are not random: hallucination targets
concentrate on a small set of high- and mid-resource {attractor} scripts, with
Latin, Arabic, and Devanagari collectively accounting for the majority of
cross-script substitutions, reflecting their dominance in OCR training corpora.
Several substitution pairs reflect genuine script-family proximity---Syriac
$\to$ Arabic, Grantha $\to$ Tamil, Coptic $\to$ Greek,
Newa $\to$ Devanagari, Tangut $\to$ Han, Lisu $\to$ Latin---pairing each lower-resource script with its closest higher-resource
relative. Other substitutions are purely distributional: Old Uyghur and
Mongolian are most often predicted as Arabic, likely because both are rendered
horizontally in our benchmark despite being traditionally vertical scripts, and
their horizontal rendering may share superficial visual features with Arabic's
connected cursive strokes. Ogham is rendered almost exclusively as Latin.
This suggests models conflate visual similarity with statistical co-occurrence
in training data, defaulting to
whichever script is most compatible with the image features they extract

\begin{mdframed}[linecolor=black!70, topline=false, bottomline=false, rightline=false, backgroundcolor=gray!0, innertopmargin=3pt, innerbottommargin=3pt]
\textbf{Finding 6.} Cross-script hallucination is the dominant failure mode: models
overwhelmingly confabulate in a wrong script rather than abstain. Hallucination
targets concentrate on high- and mid-resource attractor scripts, with some substitutions
reflecting genuine script-family proximity and visual resemblance, while others
are purely distributional, driven by the dominance of certain scripts in
OCR training corpora.
\end{mdframed}

\section{Conclusion}

We introduced \name, a comprehensive benchmark for evaluating OCR generalization
across 158 Unicode scripts, spanning clean and degraded image variants rendered
from real multilingual texts. Our evaluation of 14 open-weight and proprietary
vision-language models reveals that current systems achieve strong performance on
Latin but degrade substantially on mid-resource scripts, failing almost universally
on the remaining 148 low-resource scripts, with even the best model correctly
transcribing fewer than 7.7\% of sentences with a character error rate below 5\%
in this tier. Failure is not silent: models hallucinate fluent text in familiar
scripts rather than giving up, and script-aware hinting provides only marginal
relief in transcription accuracy, confirming that training coverage is a key
bottleneck. Our analysis shows that performance does not decline gradually from
mid- to low-resource scripts, but drops sharply, indicating that it largely depends
on whether a script is sufficiently represented during training. We release the
benchmark, pipeline, and code to support reproducible research. We hope \name\
serves as a call to action for the community to extend OCR development beyond the
handful of scripts that currently receive most attention.

%Bibliography
\bibliographystyle{plainnat}
\bibliography{main}

\clearpage
\appendix

\section{Rendering Pipeline Details}
\label{app:rendering}

\begin{figure}[t]
  \centering
  \begin{subfigure}[t]{0.48\linewidth}
    \centering
    \includegraphics[width=\linewidth]{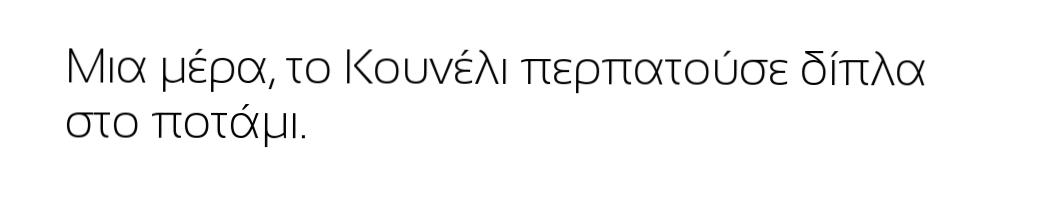}
    \caption{Greek --- clean}
    \label{fig:greek_clean}
  \end{subfigure}
  \hfill
  \begin{subfigure}[t]{0.48\linewidth}
    \centering
    \includegraphics[width=\linewidth]{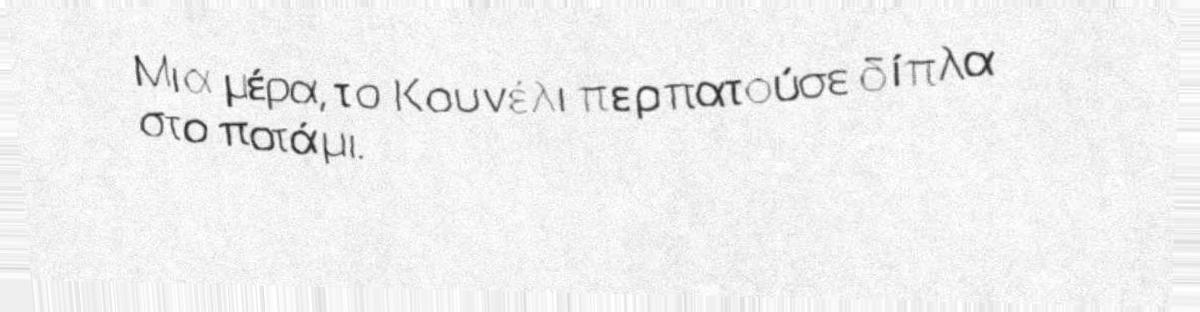}
    \caption{Greek --- degraded}
    \label{fig:greek_degraded}
  \end{subfigure}

  \vspace{0.5em}

  \begin{subfigure}[t]{0.48\linewidth}
    \centering
    \includegraphics[width=\linewidth]{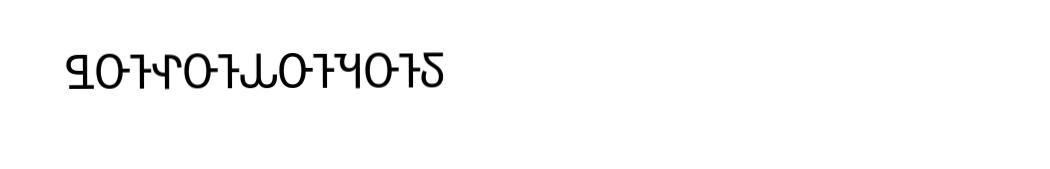}
    \caption{Aghwan --- clean}
    \label{fig:aghb_clean}
  \end{subfigure}
  \hfill
  \begin{subfigure}[t]{0.48\linewidth}
    \centering
    \includegraphics[width=\linewidth]{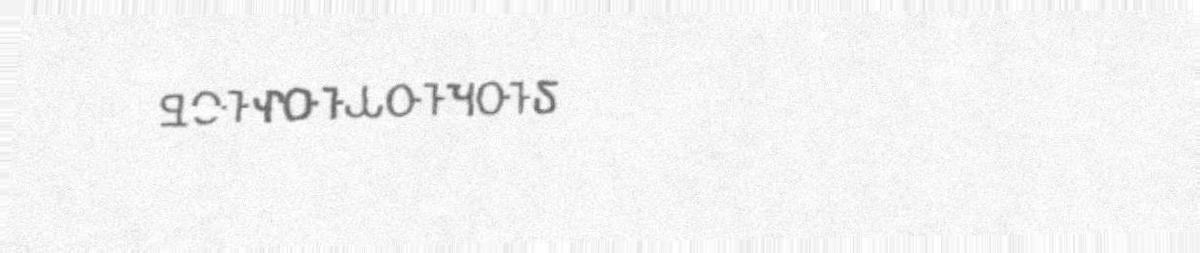}
    \caption{Aghwan --- degraded}
    \label{fig:aghb_degraded}
  \end{subfigure}

  \caption{Example images from \name for Greek (Grek, Mid tier resource) and
    Aghwan (Aghb, Low tier resource), each shown in clean and degraded variants.}
  \label{fig:sample_images}
  \vspace{10pt}
\end{figure}

Images are rendered using HarfBuzz~\citep{harfbuzz2026} for text shaping and
FreeType~\citep{freetype2024} for glyph rasterization. Sentences with mixed
bidirectional content are excluded; all rendered text is uniformly LTR or RTL.
For each sentence we produce two image variants as described below and
illustrated in Figure~\ref{fig:sample_images}.

\textbf{Clean variant.}
Text is rendered at 48px on a plain white 1000px-wide canvas with 40px padding,
with a negligible random rotation of up to $\pm 1^\circ$ to simulate minor page tilt.

\textbf{Degraded variant.}
A sequential pipeline simulates an aged physical document:

\begin{enumerate}
  \item \textbf{Paper background and rotation.} The image is placed onto a
      randomly cropped scanned paper texture, then rotated by up to
      $\pm 2^\circ$ to simulate page tilt.
  \item \textbf{Elastic deformation and Gaussian noise.} A smooth displacement
        field ($17{\times}17$ Gaussian kernel, $\pm 8$px amplitude) warps the
        image; independent Gaussian noise ($\sigma = 8$) is then added.
  \item \textbf{Ink effects.} Between 10 and 30 white rectangular patches
        (${\leq}\,40{\times}15$px) simulate ink dropout; pixel intensities are
        then scaled to 50--85\% with texture noise ($\sigma = 10$) to simulate
        ink fading.
  \item \textbf{Resolution and compression.} Images are downsampled to
        40--70\% of original resolution and upscaled back (area/bilinear
        interpolation), then JPEG-compressed at quality 30--80.
  \item \textbf{Perspective distortion.} The four corners are independently
        warped by up to 10\% of the image dimensions.
\end{enumerate}

Additionally, at the glyph level during rendering, character spacing is
perturbed by $-2$ to $+4$ pixels, each glyph is independently dilated
(prob.\ 0.4) or eroded (prob.\ 0.25) with a $2{\times}2$ kernel, each line is
vertically jittered by up to $\pm 3$ pixels, and glyphs are displaced
vertically by a parabolic curvature to simulate page curl.

These degradation operations reflect common artifacts in document capture,
scanning, and photocopying
pipelines~\citep{groleau2022augraphy}, as well as standard augmentation
practices in OCR literature~\citep{gupta2016synthetic,yim2021synthtiger}:
geometric distortions model page misalignment, photometric noise simulates
low-quality digitization, and morphological perturbations reflect ink spread
and aging. All transformations are applied within the parameter ranges listed
above, validated by human inspection across Latin, Greek, Cyrillic, and Arabic
scripts, to ensure legibility while still providing sufficient challenge for
model robustness evaluation.

\newpage

\section{Per-Script Results}
\label{app:per_script}

Table~\ref{tab:zero_sa} shows the scripts for which all models obtain zero 
in the ScriptAcc metric; for these, we report the sentence count in the 
benchmark ($n$) and the two scripts most frequently observed in model outputs 
(as a diagnostic for hallucination). Tables~\ref{tab:script_a} and 
\ref{tab:script_b} report per-script Acc@5 and ScriptAcc for all evaluated 
models, along with the sentence count and the most frequently observed output 
scripts.
\begin{table}[ht]
\centering
\caption{Scripts for which all models obtain zero in both Acc@5 and ScriptAcc. These scripts are not identifiable by any model. Top-2 Out.Script shows which 
scripts appear most frequently in model outputs for these samples (avg.\ over all models).}
\footnotesize
\label{tab:zero_sa}
\resizebox{0.9\textwidth}{!}{%
\begin{tabular}{l r p{2cm}@{\hspace{8pt}}l r p{2cm}@{\hspace{8pt}}l r p{2cm}}
\toprule
\textbf{Script} & \textbf{$n$} & \textbf{Top-2 Out.Script} & \textbf{Script} & \textbf{$n$} & \textbf{Top-2 Out.Script} & \textbf{Script} & \textbf{$n$} & \textbf{Top-2 Out.Script} \\
\midrule
Lepc & 100 & Latn, Arab & Limb & 100 & Mtei, Tibt & Mand & 100 & Arab, Syrc \\
Modi & 100 & Deva, Thai & Mong & 100 & Hani, Shaw & Newa & 100 & Deva, Beng \\
Nkoo & 100 & Arab, Latn & Ogam & 100 & Latn, Hani & Olck & 100 & Latn, Sinh \\
Hung & 100 & Latn, Cyrl & Ital & 100 & Latn, Cyrl & Ougr & 100 & Arab, Mong \\
Phli & 100 & Hebr, Latn & Prti & 100 & Hebr, Tibt & Sarb & 100 & Latn, Tibt \\
Shrd & 100 & Deva, Guru & Dsrt & 100 & Shaw, Latn & Glag & 100 & Latn, Beng \\
Gran & 100 & Taml, Thai & Ugar & 100 & Latn, Xsux & Wara & 100 & Latn, Geor \\
Bugi & 100 & Latn, Cans & Cakm & 100 & Mlym, Mymr & Cham & 100 & Mymr, Latn \\
Sund & 100 & Latn, Hebr & Tale & 100 & Latn, Hebr & Talu & 100 & Geor, Mymr \\
Tavt & 100 & Thai, Latn & Kali & 100 & Khmr, Thai & Khar & 100 & Hebr, Cans \\
Kthi & 100 & Deva, Gujr & Lana & 100 & Mymr, Tibt & Adlm & 100 & Thai, Latn \\
Ahom & 100 & Thai, Latn & Avst & 100 & Arab, Latn & Bhks & 100 & Latn, Olck \\
Sgnw & 90 & Latn, Arab & Soyo & 86 & Deva, Tibt & Saur & 85 & Latn, Beng \\
Mahj & 82 & Latn, Geor & Diak & 79 & Mlym, Beng & Zanb & 76 & Mtei, Latn \\
Tirh & 75 & Beng, Thai & Nand & 68 & Deva, Thai & Sunu & 67 & Beng, Latn \\
Mani & 65 & Arab, Latn & Hatr & 63 & Hebr, Arab & Palm & 63 & Hebr, Phnx \\
Gonm & 62 & Latn, Ethi & Nbat & 61 & Hebr, Ethi & Kits & 60 & Hani, Latn \\
Samr & 59 & Latn, Phnx & Elym & 57 & Hebr, Latn & Osge & 48 & Grek, Latn \\
Armi & 45 & Hebr, Latn & Gong & 43 & Knda, Tibt & Berf & 41 & Ethi, Latn \\
Osma & 40 & Latn, Armn & Mult & 38 & Latn, Geor & Lydi & 35 & Latn, Runr \\
Sind & 34 & Latn, Geor & Mroo & 32 & Latn, Cans & Plrd & 31 & Latn, Grek \\
Rjng & 31 & Latn, Cans & Takr & 30 & Guru, Gujr & Rohg & 28 & Arab, Latn \\
Dogr & 27 & Deva, Guru & Cprt & 27 & Latn, Tibt & Nagm & 26 & Latn, Geor \\
Kawi & 26 & Khmr, Thai & Batk & 26 & Latn, Cans & Hano & 25 & Latn, Runr \\
Lyci & 23 & Latn, Grek & Lina & 21 & Latn, Hani & Medf & 21 & Taml, Latn \\
Hluw & 21 & Latn, Egyp & Sogd & 18 & Arab, Latn & Khoj & 17 & Gujr, Beng \\
Narb & 16 & Latn, Arab & Hmnp & 16 & Thai, Latn & Toto & 16 & Geor, Grek \\
Maka & 15 & Ethi, Shaw & Phag & 15 & Latn, Beng & Wcho & 15 & Arab, Geor \\
Tnsa & 14 & Latn, Armn & Hmng & 13 & Thai, Khmr & Perm & 13 & Latn, Tibt \\
Yezi & 13 & Grek, Latn & Todr & 12 & Latn, Thai & Vith & 12 & Latn, Grek \\
Krai & 12 & Thai, Taml & Elba & 11 & Latn, Geor & Mend & 9 & Latn, Ethi \\
Cari & 9 & Latn, Grek & Buhd & 8 & Latn, Tibt & Tagb & 8 & Latn, Grek \\
Nshu & 5 & Hani, Latn & Bass & 5 & Cans, Geor & Sogo & 4 & Latn, Hani \\
Sora & 3 & Latn, Mymr & Chrs & 2 & Latn, Arab & Pauc & 1 & Thai, Cyrl \\
Phlp & 1 & Arab, Latn &  & &  &  & &  \\
\bottomrule
\end{tabular}
}%resizebox
\end{table}

% \footnotesize
% \setlength{\tabcolsep}{3pt}

\begin{table}[ht]
\centering
\caption{A@5 (Acc@5)/SA (ScriptAcc) (\%) per script. In order: PaddleOCR-VL-1.5; olmOCR-2; Gemini 3.1 Flash-Lite; dots.mocr; dots.ocr; HunyuanOCR; GPT-4.1.}
\label{tab:script_a}
\footnotesize
\resizebox{\textwidth}{!}{%
\begin{tabular}{l r p{2.2cm} r r r r r r r}
\toprule
\textbf{Script} & \textbf{$n$} & \textbf{Top-2 Out.Script} & \makecell{\textbf{Paddle-1.5}\\A@5/SA$\uparrow$} & \makecell{\textbf{olmOCR2}\\A@5/SA$\uparrow$} & \makecell{\textbf{Gemini-FL}\\A@5/SA$\uparrow$} & \makecell{\textbf{dots.mocr}\\A@5/SA$\uparrow$} & \makecell{\textbf{dots.ocr}\\A@5/SA$\uparrow$} & \makecell{\textbf{Hunyuan}\\A@5/SA$\uparrow$} & \makecell{\textbf{GPT-4.1}\\A@5/SA$\uparrow$} \\
\midrule
  Latn & 4000 & Latn & \cellcolor[HTML]{9BCCED}79.8/99.2 & \cellcolor[HTML]{8EC5EA}90.5/99.6 & \cellcolor[HTML]{8AC3EA}95.3/99.7 & \cellcolor[HTML]{8AC3E9}93.1/99.8 & \cellcolor[HTML]{8BC4EA}91.8/99.5 & \cellcolor[HTML]{95C9EC}85.3/99.5 & \cellcolor[HTML]{97CAEC}83.2/99.7 \\
  Cyrl & 400 & Cyrl, Latn & \cellcolor[HTML]{9DCDED}78.0/97.8 & \cellcolor[HTML]{A6D1EF}71.2/98.2 & \cellcolor[HTML]{92C8EB}88.8/99.2 & \cellcolor[HTML]{96C9EC}83.5/99.2 & \cellcolor[HTML]{92C8EB}86.0/99.2 & \cellcolor[HTML]{94C8EB}86.8/98.5 & \cellcolor[HTML]{A8D2EF}69.8/98.8 \\
  Hani & 400 & Hani & \cellcolor[HTML]{B3D8F1}60.8/99.2 & \cellcolor[HTML]{9DCDED}78.2/100.0 & \cellcolor[HTML]{96CAEC}85.5/99.8 & \cellcolor[HTML]{92C7EB}86.5/100.0 & \cellcolor[HTML]{98CAEC}81.5/99.2 & \cellcolor[HTML]{98CBEC}83.0/99.5 & \cellcolor[HTML]{BADCF2}55.0/99.5 \\
  Deva & 400 & Deva & \cellcolor[HTML]{FDFEFE}1.0/99.8 & \cellcolor[HTML]{A6D1EF}71.5/100.0 & \cellcolor[HTML]{8DC5EA}93.5/100.0 & \cellcolor[HTML]{92C8EB}86.0/98.5 & \cellcolor[HTML]{8DC5EA}90.2/100.0 & \cellcolor[HTML]{9CCDED}79.8/100.0 & \cellcolor[HTML]{ADD5F0}65.8/100.0 \\
  Arab & 400 & Arab, Latn & \cellcolor[HTML]{E4F1FA}21.0/99.8 & \cellcolor[HTML]{D8EBF8}31.0/99.8 & \cellcolor[HTML]{B7DAF2}58.8/99.8 & \cellcolor[HTML]{B1D7F0}62.0/97.5 & \cellcolor[HTML]{AFD6F0}63.2/98.2 & \cellcolor[HTML]{CFE7F6}38.2/97.5 & \cellcolor[HTML]{CDE6F6}39.5/99.8 \\
  Jpan & 100 & Jpan & \cellcolor[HTML]{87C2E9}96.0/100.0 & \cellcolor[HTML]{85C1E9}98.0/100.0 & \cellcolor[HTML]{87C2E9}98.0/99.0 & \cellcolor[HTML]{8BC4EA}92.0/100.0 & \cellcolor[HTML]{90C6EB}88.0/100.0 & \cellcolor[HTML]{85C1E9}99.0/100.0 & \cellcolor[HTML]{85C1E9}98.0/100.0 \\
  Hang & 100 & Hang, Latn & \cellcolor[HTML]{87C2E9}96.0/100.0 & \cellcolor[HTML]{85C1E9}98.0/100.0 & \cellcolor[HTML]{85C1E9}100.0/100.0 & \cellcolor[HTML]{9FCEED}76.0/97.0 & \cellcolor[HTML]{9FCEED}76.0/95.0 & \cellcolor[HTML]{8BC4EA}94.0/100.0 & \cellcolor[HTML]{8BC4EA}93.0/100.0 \\
  Grek & 100 & Grek, Cyrl & \cellcolor[HTML]{AAD4EF}68.0/98.0 & \cellcolor[HTML]{97CAEC}83.0/100.0 & \cellcolor[HTML]{90C6EB}91.0/100.0 & \cellcolor[HTML]{8FC6EA}89.0/100.0 & \cellcolor[HTML]{95C9EC}84.0/100.0 & \cellcolor[HTML]{A0CEED}77.0/99.0 & \cellcolor[HTML]{94C8EB}86.0/100.0 \\
  Taml & 100 & Taml, Telu & \cellcolor[HTML]{85C1E9}98.0/100.0 & \cellcolor[HTML]{FAFCFE}4.0/100.0 & \cellcolor[HTML]{86C1E9}99.0/100.0 & \cellcolor[HTML]{92C8EB}86.0/99.0 & \cellcolor[HTML]{90C6EB}88.0/100.0 & \cellcolor[HTML]{AED6F0}65.0/95.0 & \cellcolor[HTML]{A4D0EE}73.0/100.0 \\
  Telu & 100 & Telu, Beng & \cellcolor[HTML]{9ACBEC}81.0/100.0 & \cellcolor[HTML]{FFFFFF}0.0/100.0 & \cellcolor[HTML]{8EC5EA}92.0/100.0 & \cellcolor[HTML]{8DC5EA}90.0/97.0 & \cellcolor[HTML]{8AC3E9}93.0/100.0 & \cellcolor[HTML]{D3E9F7}35.0/72.0 & \cellcolor[HTML]{E6F2FA}20.0/100.0 \\
  Thai & 100 & Thai, Latn & \cellcolor[HTML]{E9F4FB}17.0/100.0 & \cellcolor[HTML]{D4E9F7}34.0/100.0 & \cellcolor[HTML]{ACD4F0}68.0/100.0 & \cellcolor[HTML]{9BCCED}79.0/99.0 & \cellcolor[HTML]{99CBEC}81.0/100.0 & \cellcolor[HTML]{AED6F0}65.0/100.0 & \cellcolor[HTML]{BFDEF3}51.0/100.0 \\
  Geor & 100 & Geor, Thai & \cellcolor[HTML]{FFFFFF}0.0/0.0 & \cellcolor[HTML]{FFFFFF}0.0/100.0 & \cellcolor[HTML]{92C7EB}89.0/97.0 & \cellcolor[HTML]{85C1E9}97.0/97.0 & \cellcolor[HTML]{85C1E9}97.0/97.0 & \cellcolor[HTML]{92C7EB}88.0/90.0 & \cellcolor[HTML]{F5F9FD}8.0/96.0 \\
  Gujr & 100 & Gujr, Deva & \cellcolor[HTML]{FFFFFF}0.0/0.0 & \cellcolor[HTML]{EEF6FC}13.0/100.0 & \cellcolor[HTML]{87C2E9}98.0/100.0 & \cellcolor[HTML]{8DC5EA}90.0/98.0 & \cellcolor[HTML]{87C2E9}95.0/100.0 & \cellcolor[HTML]{F7FBFD}6.0/9.0 & \cellcolor[HTML]{C3E0F4}48.0/100.0 \\
  Guru & 100 & Guru, Deva & \cellcolor[HTML]{FFFFFF}0.0/0.0 & \cellcolor[HTML]{FFFFFF}0.0/71.0 & \cellcolor[HTML]{97CAEC}85.0/100.0 & \cellcolor[HTML]{9BCCED}79.0/94.0 & \cellcolor[HTML]{8FC6EA}89.0/100.0 & \cellcolor[HTML]{A7D2EF}71.0/77.0 & \cellcolor[HTML]{E6F2FA}20.0/100.0 \\
  Beng & 100 & Beng, Deva & \cellcolor[HTML]{DAECF8}29.0/99.0 & \cellcolor[HTML]{DDEDF8}27.0/100.0 & \cellcolor[HTML]{B2D7F1}63.0/100.0 & \cellcolor[HTML]{B6D9F1}58.0/99.0 & \cellcolor[HTML]{A3D0EE}73.0/100.0 & \cellcolor[HTML]{BDDDF3}53.0/98.0 & \cellcolor[HTML]{DDEDF8}27.0/99.0 \\
  Tibt & 100 & Tibt, Deva & \cellcolor[HTML]{9ACBEC}81.0/100.0 & \cellcolor[HTML]{FFFFFF}0.0/86.0 & \cellcolor[HTML]{E7F3FA}19.0/100.0 & \cellcolor[HTML]{9CCDED}78.0/100.0 & \cellcolor[HTML]{9FCEED}76.0/100.0 & \cellcolor[HTML]{A7D2EF}71.0/100.0 & \cellcolor[HTML]{FFFFFF}0.0/98.0 \\
  Armn & 100 & Armn, Thai & \cellcolor[HTML]{FFFFFF}0.0/0.0 & \cellcolor[HTML]{FFFFFF}0.0/52.0 & \cellcolor[HTML]{88C2E9}97.0/100.0 & \cellcolor[HTML]{87C2E9}95.0/100.0 & \cellcolor[HTML]{8AC3E9}93.0/100.0 & \cellcolor[HTML]{EAF4FB}17.0/73.0 & \cellcolor[HTML]{FDFEFE}1.0/100.0 \\
  Knda & 100 & Knda, Telu & \cellcolor[HTML]{FFFFFF}0.0/0.0 & \cellcolor[HTML]{FFFFFF}0.0/89.0 & \cellcolor[HTML]{97CAEC}85.0/100.0 & \cellcolor[HTML]{91C7EB}87.0/89.0 & \cellcolor[HTML]{9ACBEC}80.0/90.0 & \cellcolor[HTML]{D6EAF7}33.0/76.0 & \cellcolor[HTML]{EBF4FB}16.0/94.0 \\
  Hebr & 100 & Hebr, Thai & \cellcolor[HTML]{FFFFFF}0.0/0.0 & \cellcolor[HTML]{F3F9FC}9.0/97.0 & \cellcolor[HTML]{B4D9F1}61.0/100.0 & \cellcolor[HTML]{C1DFF3}49.0/93.0 & \cellcolor[HTML]{B9DBF2}55.0/98.0 & \cellcolor[HTML]{CBE4F5}42.0/91.0 & \cellcolor[HTML]{CAE4F5}42.0/99.0 \\
  Sinh & 100 & Telu, Sinh & \cellcolor[HTML]{FFFFFF}0.0/0.0 & \cellcolor[HTML]{FFFFFF}0.0/99.0 & \cellcolor[HTML]{A4D1EE}74.0/100.0 & \cellcolor[HTML]{8FC6EA}89.0/100.0 & \cellcolor[HTML]{8FC6EA}89.0/100.0 & \cellcolor[HTML]{FCFDFE}2.0/4.0 & \cellcolor[HTML]{FDFEFE}1.0/100.0 \\
  Laoo & 100 & Laoo, Thai & \cellcolor[HTML]{FFFFFF}0.0/0.0 & \cellcolor[HTML]{FFFFFF}0.0/15.0 & \cellcolor[HTML]{A7D2EF}72.0/100.0 & \cellcolor[HTML]{A2CFEE}74.0/97.0 & \cellcolor[HTML]{9BCCED}79.0/99.0 & \cellcolor[HTML]{FFFFFF}0.0/0.0 & \cellcolor[HTML]{FFFFFF}0.0/65.0 \\
  Mlym & 100 & Mlym, Telu & \cellcolor[HTML]{FFFFFF}0.0/0.0 & \cellcolor[HTML]{FFFFFF}0.0/99.0 & \cellcolor[HTML]{9CCCED}81.0/100.0 & \cellcolor[HTML]{A8D2EF}69.0/99.0 & \cellcolor[HTML]{D3E8F7}35.0/100.0 & \cellcolor[HTML]{E2F0F9}23.0/74.0 & \cellcolor[HTML]{ECF5FB}15.0/100.0 \\
  Ethi & 100 & Ethi, Tibt & \cellcolor[HTML]{FFFFFF}0.0/0.0 & \cellcolor[HTML]{FFFFFF}0.0/97.0 & \cellcolor[HTML]{E7F3FA}19.0/100.0 & \cellcolor[HTML]{BEDEF3}51.0/96.0 & \cellcolor[HTML]{ADD5F0}65.0/100.0 & \cellcolor[HTML]{E8F3FB}18.0/29.0 & \cellcolor[HTML]{FFFFFF}0.0/100.0 \\
  Thaa & 100 & Thaa, Arab & \cellcolor[HTML]{FFFFFF}0.0/0.0 & \cellcolor[HTML]{FFFFFF}0.0/0.0 & \cellcolor[HTML]{FFFFFF}0.0/100.0 & \cellcolor[HTML]{B6D9F1}58.0/98.0 & \cellcolor[HTML]{A8D2EF}69.0/100.0 & \cellcolor[HTML]{FFFFFF}0.0/0.0 & \cellcolor[HTML]{FFFFFF}0.0/58.0 \\
  Orya & 100 & Orya, Thai & \cellcolor[HTML]{FFFFFF}0.0/0.0 & \cellcolor[HTML]{FFFFFF}0.0/91.0 & \cellcolor[HTML]{9BCCED}82.0/100.0 & \cellcolor[HTML]{F1F7FC}11.0/93.0 & \cellcolor[HTML]{FFFFFF}0.0/4.0 & \cellcolor[HTML]{FCFDFE}2.0/3.0 & \cellcolor[HTML]{FAFCFE}4.0/100.0 \\
  Mymr & 100 & Mymr, Thai & \cellcolor[HTML]{FFFFFF}0.0/0.0 & \cellcolor[HTML]{FFFFFF}0.0/75.0 & \cellcolor[HTML]{D7EBF7}32.0/99.0 & \cellcolor[HTML]{ECF5FB}15.0/95.0 & \cellcolor[HTML]{EEF6FC}13.0/94.0 & \cellcolor[HTML]{EBF4FB}16.0/62.0 & \cellcolor[HTML]{FFFFFF}0.0/99.0 \\
  Khmr & 100 & Khmr, Thai & \cellcolor[HTML]{FFFFFF}0.0/0.0 & \cellcolor[HTML]{FFFFFF}0.0/96.0 & \cellcolor[HTML]{DCEDF8}28.0/100.0 & \cellcolor[HTML]{F4F9FD}8.0/95.0 & \cellcolor[HTML]{F7FBFD}6.0/89.0 & \cellcolor[HTML]{FDFEFE}1.0/43.0 & \cellcolor[HTML]{FFFFFF}0.0/100.0 \\
  Copt & 100 & Grek, Copt & \cellcolor[HTML]{FFFFFF}0.0/0.0 & \cellcolor[HTML]{FFFFFF}0.0/0.0 & \cellcolor[HTML]{E4F1FA}22.0/36.0 & \cellcolor[HTML]{FFFFFF}0.0/0.0 & \cellcolor[HTML]{FFFFFF}0.0/0.0 & \cellcolor[HTML]{FFFFFF}0.0/0.0 & \cellcolor[HTML]{FDFEFE}1.0/31.0 \\
  Bopo & 100 & Bopo, Hani & \cellcolor[HTML]{FBFDFE}3.0/20.0 & \cellcolor[HTML]{FFFFFF}0.0/13.0 & \cellcolor[HTML]{FDFEFE}1.0/97.0 & \cellcolor[HTML]{FFFFFF}0.0/19.0 & \cellcolor[HTML]{FFFFFF}0.0/41.0 & \cellcolor[HTML]{FFFFFF}0.0/24.0 & \cellcolor[HTML]{FFFFFF}0.0/0.0 \\
  Cans & 100 & Cans, Latn & \cellcolor[HTML]{FFFFFF}0.0/0.0 & \cellcolor[HTML]{FFFFFF}0.0/0.0 & \cellcolor[HTML]{FFFFFF}0.0/74.0 & \cellcolor[HTML]{FFFFFF}0.0/0.0 & \cellcolor[HTML]{FFFFFF}0.0/1.0 & \cellcolor[HTML]{FFFFFF}0.0/0.0 & \cellcolor[HTML]{FFFFFF}0.0/97.0 \\
  Egyp & 100 & Egyp, Latn & \cellcolor[HTML]{FFFFFF}0.0/0.0 & \cellcolor[HTML]{FFFFFF}0.0/0.0 & \cellcolor[HTML]{FFFFFF}0.0/99.0 & \cellcolor[HTML]{FFFFFF}0.0/0.0 & \cellcolor[HTML]{FFFFFF}0.0/0.0 & \cellcolor[HTML]{FFFFFF}0.0/0.0 & \cellcolor[HTML]{FFFFFF}0.0/56.0 \\
  Xsux & 100 & Xsux, Hani & \cellcolor[HTML]{FFFFFF}0.0/0.0 & \cellcolor[HTML]{FFFFFF}0.0/0.0 & \cellcolor[HTML]{FFFFFF}0.0/71.0 & \cellcolor[HTML]{FFFFFF}0.0/0.0 & \cellcolor[HTML]{FFFFFF}0.0/0.0 & \cellcolor[HTML]{FFFFFF}0.0/0.0 & \cellcolor[HTML]{FFFFFF}0.0/54.0 \\
  Syrc & 100 & Arab, Syrc & \cellcolor[HTML]{FFFFFF}0.0/0.0 & \cellcolor[HTML]{FFFFFF}0.0/0.0 & \cellcolor[HTML]{FFFFFF}0.0/84.0 & \cellcolor[HTML]{FFFFFF}0.0/0.0 & \cellcolor[HTML]{FFFFFF}0.0/0.0 & \cellcolor[HTML]{FFFFFF}0.0/0.0 & \cellcolor[HTML]{FFFFFF}0.0/20.0 \\
  Runr & 100 & Runr, Latn & \cellcolor[HTML]{FFFFFF}0.0/0.0 & \cellcolor[HTML]{FFFFFF}0.0/0.0 & \cellcolor[HTML]{FFFFFF}0.0/84.0 & \cellcolor[HTML]{FFFFFF}0.0/0.0 & \cellcolor[HTML]{FFFFFF}0.0/0.0 & \cellcolor[HTML]{FFFFFF}0.0/0.0 & \cellcolor[HTML]{FFFFFF}0.0/16.0 \\
  Mtei & 100 & Mtei, Tibt & \cellcolor[HTML]{FFFFFF}0.0/0.0 & \cellcolor[HTML]{FFFFFF}0.0/0.0 & \cellcolor[HTML]{FFFFFF}0.0/97.0 & \cellcolor[HTML]{FFFFFF}0.0/0.0 & \cellcolor[HTML]{FFFFFF}0.0/0.0 & \cellcolor[HTML]{FFFFFF}0.0/0.0 & \cellcolor[HTML]{FFFFFF}0.0/0.0 \\
  Brai & 100 & Brai, Latn & \cellcolor[HTML]{FFFFFF}0.0/0.0 & \cellcolor[HTML]{FFFFFF}0.0/0.0 & \cellcolor[HTML]{FFFFFF}0.0/88.0 & \cellcolor[HTML]{FFFFFF}0.0/0.0 & \cellcolor[HTML]{FFFFFF}0.0/0.0 & \cellcolor[HTML]{FFFFFF}0.0/0.0 & \cellcolor[HTML]{FFFFFF}0.0/0.0 \\
  Cher & 100 & Cher, Latn & \cellcolor[HTML]{FFFFFF}0.0/0.0 & \cellcolor[HTML]{FFFFFF}0.0/0.0 & \cellcolor[HTML]{FFFFFF}0.0/88.0 & \cellcolor[HTML]{FFFFFF}0.0/0.0 & \cellcolor[HTML]{FFFFFF}0.0/0.0 & \cellcolor[HTML]{FFFFFF}0.0/0.0 & \cellcolor[HTML]{FFFFFF}0.0/0.0 \\
  Xpeo & 100 & Xpeo, Latn & \cellcolor[HTML]{FFFFFF}0.0/0.0 & \cellcolor[HTML]{FFFFFF}0.0/0.0 & \cellcolor[HTML]{FFFFFF}0.0/77.0 & \cellcolor[HTML]{FFFFFF}0.0/0.0 & \cellcolor[HTML]{FFFFFF}0.0/0.0 & \cellcolor[HTML]{FFFFFF}0.0/0.0 & \cellcolor[HTML]{FFFFFF}0.0/0.0 \\
  Tglg & 100 & Geor, Shaw & \cellcolor[HTML]{FFFFFF}0.0/0.0 & \cellcolor[HTML]{FFFFFF}0.0/0.0 & \cellcolor[HTML]{FFFFFF}0.0/25.0 & \cellcolor[HTML]{FFFFFF}0.0/0.0 & \cellcolor[HTML]{FFFFFF}0.0/0.0 & \cellcolor[HTML]{FFFFFF}0.0/0.0 & \cellcolor[HTML]{FFFFFF}0.0/0.0 \\
  Java & 100 & Khmr, Mlym & \cellcolor[HTML]{FFFFFF}0.0/0.0 & \cellcolor[HTML]{FFFFFF}0.0/0.0 & \cellcolor[HTML]{FFFFFF}0.0/15.0 & \cellcolor[HTML]{FFFFFF}0.0/0.0 & \cellcolor[HTML]{FFFFFF}0.0/0.0 & \cellcolor[HTML]{FFFFFF}0.0/0.0 & \cellcolor[HTML]{FFFFFF}0.0/0.0 \\
  Bali & 100 & Khmr, Laoo & \cellcolor[HTML]{FFFFFF}0.0/0.0 & \cellcolor[HTML]{FFFFFF}0.0/0.0 & \cellcolor[HTML]{FFFFFF}0.0/14.0 & \cellcolor[HTML]{FFFFFF}0.0/0.0 & \cellcolor[HTML]{FFFFFF}0.0/0.0 & \cellcolor[HTML]{FFFFFF}0.0/0.0 & \cellcolor[HTML]{FFFFFF}0.0/0.0 \\
  Phnx & 100 & Latn, Tibt & \cellcolor[HTML]{FFFFFF}0.0/0.0 & \cellcolor[HTML]{FFFFFF}0.0/0.0 & \cellcolor[HTML]{FFFFFF}0.0/8.0 & \cellcolor[HTML]{FFFFFF}0.0/0.0 & \cellcolor[HTML]{FFFFFF}0.0/0.0 & \cellcolor[HTML]{FFFFFF}0.0/0.0 & \cellcolor[HTML]{FFFFFF}0.0/0.0 \\
  Linb & 100 & Latn, Hani & \cellcolor[HTML]{FFFFFF}0.0/0.0 & \cellcolor[HTML]{FFFFFF}0.0/0.0 & \cellcolor[HTML]{FFFFFF}0.0/7.0 & \cellcolor[HTML]{FFFFFF}0.0/0.0 & \cellcolor[HTML]{FFFFFF}0.0/0.0 & \cellcolor[HTML]{FFFFFF}0.0/0.0 & \cellcolor[HTML]{FFFFFF}0.0/0.0 \\
  Vaii & 100 & Shaw, Ethi & \cellcolor[HTML]{FFFFFF}0.0/0.0 & \cellcolor[HTML]{FFFFFF}0.0/0.0 & \cellcolor[HTML]{FFFFFF}0.0/4.0 & \cellcolor[HTML]{FFFFFF}0.0/0.0 & \cellcolor[HTML]{FFFFFF}0.0/0.0 & \cellcolor[HTML]{FFFFFF}0.0/0.0 & \cellcolor[HTML]{FFFFFF}0.0/0.0 \\
  Tang & 100 & Hani, Latn & \cellcolor[HTML]{FFFFFF}0.0/0.0 & \cellcolor[HTML]{FFFFFF}0.0/0.0 & \cellcolor[HTML]{FFFFFF}0.0/3.0 & \cellcolor[HTML]{FFFFFF}0.0/0.0 & \cellcolor[HTML]{FFFFFF}0.0/0.0 & \cellcolor[HTML]{FFFFFF}0.0/0.0 & \cellcolor[HTML]{FFFFFF}0.0/0.0 \\
  Orkh & 100 & Runr, Latn & \cellcolor[HTML]{FFFFFF}0.0/0.0 & \cellcolor[HTML]{FFFFFF}0.0/0.0 & \cellcolor[HTML]{FFFFFF}0.0/0.0 & \cellcolor[HTML]{FFFFFF}0.0/0.0 & \cellcolor[HTML]{FFFFFF}0.0/0.0 & \cellcolor[HTML]{FFFFFF}0.0/0.0 & \cellcolor[HTML]{FFFFFF}0.0/2.0 \\
  Brah & 100 & Cans, Ethi & \cellcolor[HTML]{FFFFFF}0.0/0.0 & \cellcolor[HTML]{FFFFFF}0.0/0.0 & \cellcolor[HTML]{FFFFFF}0.0/2.0 & \cellcolor[HTML]{FFFFFF}0.0/0.0 & \cellcolor[HTML]{FFFFFF}0.0/0.0 & \cellcolor[HTML]{FFFFFF}0.0/0.0 & \cellcolor[HTML]{FFFFFF}0.0/0.0 \\
  Lisu & 100 & Latn, Grek & \cellcolor[HTML]{FFFFFF}0.0/0.0 & \cellcolor[HTML]{FFFFFF}0.0/0.0 & \cellcolor[HTML]{FFFFFF}0.0/1.0 & \cellcolor[HTML]{FFFFFF}0.0/0.0 & \cellcolor[HTML]{FFFFFF}0.0/0.0 & \cellcolor[HTML]{FFFFFF}0.0/0.0 & \cellcolor[HTML]{FFFFFF}0.0/0.0 \\
  Shaw & 100 & Latn, Arab & \cellcolor[HTML]{FFFFFF}0.0/0.0 & \cellcolor[HTML]{FFFFFF}0.0/0.0 & \cellcolor[HTML]{FFFFFF}0.0/1.0 & \cellcolor[HTML]{FFFFFF}0.0/0.0 & \cellcolor[HTML]{FFFFFF}0.0/0.0 & \cellcolor[HTML]{FFFFFF}0.0/0.0 & \cellcolor[HTML]{FFFFFF}0.0/0.0 \\
  Goth & 100 & Latn, Grek & \cellcolor[HTML]{FFFFFF}0.0/0.0 & \cellcolor[HTML]{FFFFFF}0.0/0.0 & \cellcolor[HTML]{FFFFFF}0.0/1.0 & \cellcolor[HTML]{FFFFFF}0.0/0.0 & \cellcolor[HTML]{FFFFFF}0.0/0.0 & \cellcolor[HTML]{FFFFFF}0.0/0.0 & \cellcolor[HTML]{FFFFFF}0.0/0.0 \\
  Tfng & 100 & Latn, Grek & \cellcolor[HTML]{FFFFFF}0.0/0.0 & \cellcolor[HTML]{FFFFFF}0.0/0.0 & \cellcolor[HTML]{FFFFFF}0.0/1.0 & \cellcolor[HTML]{FFFFFF}0.0/0.0 & \cellcolor[HTML]{FFFFFF}0.0/0.0 & \cellcolor[HTML]{FFFFFF}0.0/0.0 & \cellcolor[HTML]{FFFFFF}0.0/0.0 \\
  Yiii & 100 & Hani, Hang & \cellcolor[HTML]{FFFFFF}0.0/0.0 & \cellcolor[HTML]{FFFFFF}0.0/0.0 & \cellcolor[HTML]{FFFFFF}0.0/0.0 & \cellcolor[HTML]{FFFFFF}0.0/0.0 & \cellcolor[HTML]{FFFFFF}0.0/0.0 & \cellcolor[HTML]{FFFFFF}0.0/0.0 & \cellcolor[HTML]{FFFFFF}0.0/1.0 \\
  Sylo & 100 & Deva, Beng & \cellcolor[HTML]{FFFFFF}0.0/0.0 & \cellcolor[HTML]{FFFFFF}0.0/0.0 & \cellcolor[HTML]{FFFFFF}0.0/1.0 & \cellcolor[HTML]{FFFFFF}0.0/0.0 & \cellcolor[HTML]{FFFFFF}0.0/0.0 & \cellcolor[HTML]{FFFFFF}0.0/0.0 & \cellcolor[HTML]{FFFFFF}0.0/0.0 \\
  Aghb & 100 & Latn, Armn & \cellcolor[HTML]{FFFFFF}0.0/0.0 & \cellcolor[HTML]{FFFFFF}0.0/0.0 & \cellcolor[HTML]{FFFFFF}0.0/1.0 & \cellcolor[HTML]{FFFFFF}0.0/0.0 & \cellcolor[HTML]{FFFFFF}0.0/0.0 & \cellcolor[HTML]{FFFFFF}0.0/0.0 & \cellcolor[HTML]{FFFFFF}0.0/0.0 \\
  Sidd & 75 & Deva, Tibt & \cellcolor[HTML]{FFFFFF}0.0/0.0 & \cellcolor[HTML]{FFFFFF}0.0/0.0 & \cellcolor[HTML]{FFFFFF}0.0/0.0 & \cellcolor[HTML]{FFFFFF}0.0/0.0 & \cellcolor[HTML]{FFFFFF}0.0/0.0 & \cellcolor[HTML]{FFFFFF}0.0/0.0 & \cellcolor[HTML]{FFFFFF}0.0/1.3 \\
\bottomrule
\end{tabular}
}%resizebox
\end{table}

\begin{table}[ht]
\centering
\caption[]{A@5 (Acc@5)/SA (ScriptAcc) (\%) per script. In order: Qwen3-VL-8B; GLM-OCR; RolmOCR; LightOnOCR-2; DeepSeek-OCR-2; FireRed-OCR; Nanonets-OCR2.}
\label{tab:script_b}
\footnotesize
\resizebox{\textwidth}{!}{%
\begin{tabular}{l r p{2.2cm} r r r r r r r}
\toprule
\textbf{Script} & \textbf{$n$} & \textbf{Top-2 Out.Script} & \makecell{\textbf{Qwen3-8B}\\A@5/SA$\uparrow$} & \makecell{\textbf{GLM}\\A@5/SA$\uparrow$} & \makecell{\textbf{Rolm}\\A@5/SA$\uparrow$} & \makecell{\textbf{LightOn-2}\\A@5/SA$\uparrow$} & \makecell{\textbf{DeepSeek-2}\\A@5/SA$\uparrow$} & \makecell{\textbf{FireRed}\\A@5/SA$\uparrow$} & \makecell{\textbf{Nanonets-2}\\A@5/SA$\uparrow$} \\
\midrule
  Latn & 4000 & Latn & \cellcolor[HTML]{8EC5EA}89.5/99.7 & \cellcolor[HTML]{88C2E9}89.5/99.6 & \cellcolor[HTML]{8EC5EA}89.6/99.8 & \cellcolor[HTML]{85C1E9}89.8/99.8 & \cellcolor[HTML]{8AC3EA}76.2/98.2 & \cellcolor[HTML]{93C8EB}83.6/99.5 & \cellcolor[HTML]{89C3E9}88.6/99.8 \\
  Cyrl & 400 & Cyrl, Latn & \cellcolor[HTML]{9BCCED}79.0/99.2 & \cellcolor[HTML]{DCEDF8}26.2/98.2 & \cellcolor[HTML]{A0CEED}75.2/98.8 & \cellcolor[HTML]{A1CFEE}68.8/95.5 & \cellcolor[HTML]{A7D2EF}57.8/94.2 & \cellcolor[HTML]{B2D8F1}59.2/95.0 & \cellcolor[HTML]{AAD4EF}63.7/98.2 \\
  Hani & 400 & Hani, Latn & \cellcolor[HTML]{9CCDED}78.0/100.0 & \cellcolor[HTML]{92C7EB}82.0/100.0 & \cellcolor[HTML]{A8D3EF}68.8/99.8 & \cellcolor[HTML]{DBECF8}26.2/98.8 & \cellcolor[HTML]{B5D9F1}48.5/100.0 & \cellcolor[HTML]{A9D3EF}66.8/99.5 & \cellcolor[HTML]{C1DFF3}46.5/100.0 \\
  Deva & 400 & Deva, Guru & \cellcolor[HTML]{A6D1EF}70.8/100.0 & \cellcolor[HTML]{FEFEFE}0.2/100.0 & \cellcolor[HTML]{A6D2EF}70.2/100.0 & \cellcolor[HTML]{A6D1EF}65.2/96.0 & \cellcolor[HTML]{D3E8F7}28.5/93.8 & \cellcolor[HTML]{C3E0F4}46.5/99.0 & \cellcolor[HTML]{B7DAF2}54.2/99.8 \\
  Arab & 400 & Arab, Latn & \cellcolor[HTML]{D0E7F6}36.8/99.8 & \cellcolor[HTML]{FEFEFE}0.2/99.8 & \cellcolor[HTML]{D5EAF7}32.8/99.8 & \cellcolor[HTML]{D6EAF7}30.0/99.2 & \cellcolor[HTML]{D3E8F7}28.5/99.0 & \cellcolor[HTML]{EEF6FB}13.0/89.8 & \cellcolor[HTML]{D8EBF8}29.2/99.8 \\
  Jpan & 100 & Jpan & \cellcolor[HTML]{87C2E9}95.0/100.0 & \cellcolor[HTML]{85C1E9}92.0/100.0 & \cellcolor[HTML]{8AC3E9}93.0/100.0 & \cellcolor[HTML]{8EC5EA}83.0/100.0 & \cellcolor[HTML]{85C1E9}80.0/100.0 & \cellcolor[HTML]{85C1E9}95.0/100.0 & \cellcolor[HTML]{85C1E9}92.0/100.0 \\
  Hang & 100 & Hang, Latn & \cellcolor[HTML]{85C1E9}97.0/100.0 & \cellcolor[HTML]{E8F3FA}17.0/100.0 & \cellcolor[HTML]{85C1E9}97.0/100.0 & \cellcolor[HTML]{B7DAF2}53.0/93.0 & \cellcolor[HTML]{CFE6F6}31.0/84.0 & \cellcolor[HTML]{ABD4EF}65.0/95.0 & \cellcolor[HTML]{93C8EB}81.0/100.0 \\
  Grek & 100 & Grek, Latn & \cellcolor[HTML]{96C9EC}83.0/100.0 & \cellcolor[HTML]{BCDDF3}50.0/94.0 & \cellcolor[HTML]{97CAEC}82.0/100.0 & \cellcolor[HTML]{99CBEC}75.0/100.0 & \cellcolor[HTML]{92C8EB}71.0/100.0 & \cellcolor[HTML]{B5D9F1}57.0/93.0 & \cellcolor[HTML]{A2CFEE}70.0/100.0 \\
  Taml & 100 & Taml, Thai & \cellcolor[HTML]{F6FAFD}7.0/100.0 & \cellcolor[HTML]{FFFFFF}0.0/100.0 & \cellcolor[HTML]{FBFDFE}3.0/100.0 & \cellcolor[HTML]{D7EAF7}29.0/88.0 & \cellcolor[HTML]{D5EAF7}27.0/62.0 & \cellcolor[HTML]{FDFEFE}1.0/93.0 & \cellcolor[HTML]{FFFFFF}0.0/100.0 \\
  Telu & 100 & Telu, Knda & \cellcolor[HTML]{FFFFFF}0.0/98.0 & \cellcolor[HTML]{FFFFFF}0.0/1.0 & \cellcolor[HTML]{FFFFFF}0.0/100.0 & \cellcolor[HTML]{FFFFFF}0.0/31.0 & \cellcolor[HTML]{FDFEFE}1.0/20.0 & \cellcolor[HTML]{FFFFFF}0.0/94.0 & \cellcolor[HTML]{FFFFFF}0.0/98.0 \\
  Thai & 100 & Thai, Tibt & \cellcolor[HTML]{D8EBF7}31.0/100.0 & \cellcolor[HTML]{FFFFFF}0.0/94.0 & \cellcolor[HTML]{D5E9F7}33.0/100.0 & \cellcolor[HTML]{DDEDF8}25.0/98.0 & \cellcolor[HTML]{F4F9FD}7.0/74.0 & \cellcolor[HTML]{EDF5FB}14.0/87.0 & \cellcolor[HTML]{E1F0F9}22.0/100.0 \\
  Geor & 100 & Geor, Thai & \cellcolor[HTML]{E9F4FB}17.0/97.0 & \cellcolor[HTML]{FFFFFF}0.0/99.0 & \cellcolor[HTML]{FFFFFF}0.0/97.0 & \cellcolor[HTML]{CEE6F6}36.0/46.0 & \cellcolor[HTML]{FDFEFE}1.0/3.0 & \cellcolor[HTML]{FFFFFF}0.0/16.0 & \cellcolor[HTML]{FFFFFF}0.0/97.0 \\
  Gujr & 100 & Deva, Gujr & \cellcolor[HTML]{EEF6FC}13.0/46.0 & \cellcolor[HTML]{FFFFFF}0.0/0.0 & \cellcolor[HTML]{F9FCFE}4.0/71.0 & \cellcolor[HTML]{F6FAFD}6.0/8.0 & \cellcolor[HTML]{F5FAFD}6.0/11.0 & \cellcolor[HTML]{FFFFFF}0.0/0.0 & \cellcolor[HTML]{FFFFFF}0.0/18.0 \\
  Guru & 100 & Deva, Guru & \cellcolor[HTML]{E3F0FA}22.0/85.0 & \cellcolor[HTML]{FFFFFF}0.0/0.0 & \cellcolor[HTML]{FFFFFF}0.0/28.0 & \cellcolor[HTML]{FCFDFE}2.0/27.0 & \cellcolor[HTML]{FFFFFF}0.0/0.0 & \cellcolor[HTML]{FCFDFE}2.0/43.0 & \cellcolor[HTML]{FFFFFF}0.0/1.0 \\
  Beng & 100 & Beng, Deva & \cellcolor[HTML]{D5E9F7}33.0/100.0 & \cellcolor[HTML]{FFFFFF}0.0/96.0 & \cellcolor[HTML]{D9EBF8}30.0/100.0 & \cellcolor[HTML]{EAF4FB}15.0/83.0 & \cellcolor[HTML]{FFFFFF}0.0/22.0 & \cellcolor[HTML]{E5F1FA}20.0/100.0 & \cellcolor[HTML]{DCEDF8}26.0/100.0 \\
  Tibt & 100 & Tibt, Beng & \cellcolor[HTML]{EFF7FC}12.0/98.0 & \cellcolor[HTML]{FFFFFF}0.0/97.0 & \cellcolor[HTML]{FFFFFF}0.0/44.0 & \cellcolor[HTML]{E2F0F9}21.0/82.0 & \cellcolor[HTML]{FDFEFE}1.0/70.0 & \cellcolor[HTML]{FCFDFE}2.0/60.0 & \cellcolor[HTML]{FFFFFF}0.0/10.0 \\
  Armn & 100 & Armn, Latn & \cellcolor[HTML]{FFFFFF}0.0/48.0 & \cellcolor[HTML]{FFFFFF}0.0/9.0 & \cellcolor[HTML]{FFFFFF}0.0/98.0 & \cellcolor[HTML]{FFFFFF}0.0/13.0 & \cellcolor[HTML]{FBFDFE}2.0/11.0 & \cellcolor[HTML]{FFFFFF}0.0/3.0 & \cellcolor[HTML]{FFFFFF}0.0/87.0 \\
  Knda & 100 & Knda, Telu & \cellcolor[HTML]{FFFFFF}0.0/5.0 & \cellcolor[HTML]{FFFFFF}0.0/79.0 & \cellcolor[HTML]{FFFFFF}0.0/74.0 & \cellcolor[HTML]{FFFFFF}0.0/7.0 & \cellcolor[HTML]{FFFFFF}0.0/0.0 & \cellcolor[HTML]{FFFFFF}0.0/0.0 & \cellcolor[HTML]{FFFFFF}0.0/77.0 \\
  Hebr & 100 & Hebr, Thai & \cellcolor[HTML]{D5E9F7}33.0/99.0 & \cellcolor[HTML]{FFFFFF}0.0/95.0 & \cellcolor[HTML]{F9FCFE}4.0/93.0 & \cellcolor[HTML]{CBE4F5}38.0/95.0 & \cellcolor[HTML]{F7FBFD}5.0/31.0 & \cellcolor[HTML]{FCFDFE}2.0/64.0 & \cellcolor[HTML]{FDFEFE}1.0/92.0 \\
  Sinh & 100 & Sinh, Tibt & \cellcolor[HTML]{FFFFFF}0.0/26.0 & \cellcolor[HTML]{FFFFFF}0.0/63.0 & \cellcolor[HTML]{FFFFFF}0.0/59.0 & \cellcolor[HTML]{FFFFFF}0.0/4.0 & \cellcolor[HTML]{FFFFFF}0.0/1.0 & \cellcolor[HTML]{FFFFFF}0.0/0.0 & \cellcolor[HTML]{FFFFFF}0.0/43.0 \\
  Laoo & 100 & Thai, Laoo & \cellcolor[HTML]{F1F7FC}11.0/49.0 & \cellcolor[HTML]{FFFFFF}0.0/0.0 & \cellcolor[HTML]{FFFFFF}0.0/0.0 & \cellcolor[HTML]{FFFFFF}0.0/6.0 & \cellcolor[HTML]{FDFEFE}1.0/23.0 & \cellcolor[HTML]{FDFEFE}1.0/51.0 & \cellcolor[HTML]{FFFFFF}0.0/0.0 \\
  Mlym & 100 & Mlym, Thai & \cellcolor[HTML]{FFFFFF}0.0/80.0 & \cellcolor[HTML]{FFFFFF}0.0/100.0 & \cellcolor[HTML]{FFFFFF}0.0/99.0 & \cellcolor[HTML]{FFFFFF}0.0/9.0 & \cellcolor[HTML]{FFFFFF}0.0/0.0 & \cellcolor[HTML]{FFFFFF}0.0/0.0 & \cellcolor[HTML]{FFFFFF}0.0/92.0 \\
  Ethi & 100 & Mymr, Latn & \cellcolor[HTML]{FFFFFF}0.0/16.0 & \cellcolor[HTML]{FFFFFF}0.0/0.0 & \cellcolor[HTML]{FFFFFF}0.0/89.0 & \cellcolor[HTML]{FFFFFF}0.0/1.0 & \cellcolor[HTML]{FFFFFF}0.0/3.0 & \cellcolor[HTML]{FFFFFF}0.0/0.0 & \cellcolor[HTML]{FFFFFF}0.0/4.0 \\
  Thaa & 100 & Arab, Latn & \cellcolor[HTML]{FFFFFF}0.0/0.0 & \cellcolor[HTML]{FFFFFF}0.0/0.0 & \cellcolor[HTML]{FFFFFF}0.0/0.0 & \cellcolor[HTML]{FFFFFF}0.0/0.0 & \cellcolor[HTML]{FFFFFF}0.0/4.0 & \cellcolor[HTML]{FFFFFF}0.0/0.0 & \cellcolor[HTML]{FFFFFF}0.0/0.0 \\
  Orya & 100 & Beng, Orya & \cellcolor[HTML]{FFFFFF}0.0/17.0 & \cellcolor[HTML]{FFFFFF}0.0/1.0 & \cellcolor[HTML]{FFFFFF}0.0/74.0 & \cellcolor[HTML]{FFFFFF}0.0/0.0 & \cellcolor[HTML]{FFFFFF}0.0/0.0 & \cellcolor[HTML]{FFFFFF}0.0/0.0 & \cellcolor[HTML]{FFFFFF}0.0/9.0 \\
  Mymr & 100 & Mymr, Latn & \cellcolor[HTML]{FFFFFF}0.0/62.0 & \cellcolor[HTML]{FFFFFF}0.0/85.0 & \cellcolor[HTML]{FFFFFF}0.0/85.0 & \cellcolor[HTML]{FFFFFF}0.0/14.0 & \cellcolor[HTML]{FFFFFF}0.0/15.0 & \cellcolor[HTML]{FFFFFF}0.0/2.0 & \cellcolor[HTML]{FFFFFF}0.0/64.0 \\
  Khmr & 100 & Khmr, Thai & \cellcolor[HTML]{FFFFFF}0.0/21.0 & \cellcolor[HTML]{FFFFFF}0.0/89.0 & \cellcolor[HTML]{FFFFFF}0.0/79.0 & \cellcolor[HTML]{FFFFFF}0.0/18.0 & \cellcolor[HTML]{FFFFFF}0.0/10.0 & \cellcolor[HTML]{FFFFFF}0.0/0.0 & \cellcolor[HTML]{FFFFFF}0.0/36.0 \\
  Copt & 100 & Grek, Cyrl & \cellcolor[HTML]{FFFFFF}0.0/0.0 & \cellcolor[HTML]{FFFFFF}0.0/0.0 & \cellcolor[HTML]{FFFFFF}0.0/0.0 & \cellcolor[HTML]{FFFFFF}0.0/0.0 & \cellcolor[HTML]{FFFFFF}0.0/0.0 & \cellcolor[HTML]{FFFFFF}0.0/0.0 & \cellcolor[HTML]{FFFFFF}0.0/0.0 \\
  Bopo & 100 & Kana, Hani & \cellcolor[HTML]{FFFFFF}0.0/11.0 & \cellcolor[HTML]{FFFFFF}0.0/5.0 & \cellcolor[HTML]{FFFFFF}0.0/16.0 & \cellcolor[HTML]{FFFFFF}0.0/0.0 & \cellcolor[HTML]{FFFFFF}0.0/0.0 & \cellcolor[HTML]{FFFFFF}0.0/34.0 & \cellcolor[HTML]{FFFFFF}0.0/20.0 \\
  Cans & 100 & Latn, Grek & \cellcolor[HTML]{FFFFFF}0.0/0.0 & \cellcolor[HTML]{FFFFFF}0.0/0.0 & \cellcolor[HTML]{FFFFFF}0.0/0.0 & \cellcolor[HTML]{FFFFFF}0.0/0.0 & \cellcolor[HTML]{FFFFFF}0.0/0.0 & \cellcolor[HTML]{FFFFFF}0.0/0.0 & \cellcolor[HTML]{FFFFFF}0.0/0.0 \\
  Egyp & 100 & Latn, Avst & \cellcolor[HTML]{FFFFFF}0.0/2.0 & \cellcolor[HTML]{FFFFFF}0.0/0.0 & \cellcolor[HTML]{FFFFFF}0.0/0.0 & \cellcolor[HTML]{FFFFFF}0.0/0.0 & \cellcolor[HTML]{FFFFFF}0.0/0.0 & \cellcolor[HTML]{FFFFFF}0.0/0.0 & \cellcolor[HTML]{FFFFFF}0.0/1.0 \\
  Xsux & 100 & Latn, Hani & \cellcolor[HTML]{FFFFFF}0.0/0.0 & \cellcolor[HTML]{FFFFFF}0.0/0.0 & \cellcolor[HTML]{FFFFFF}0.0/5.0 & \cellcolor[HTML]{FFFFFF}0.0/0.0 & \cellcolor[HTML]{FFFFFF}0.0/0.0 & \cellcolor[HTML]{FFFFFF}0.0/0.0 & \cellcolor[HTML]{FFFFFF}0.0/0.0 \\
  Syrc & 100 & Arab, Hebr & \cellcolor[HTML]{FFFFFF}0.0/0.0 & \cellcolor[HTML]{FFFFFF}0.0/0.0 & \cellcolor[HTML]{FFFFFF}0.0/0.0 & \cellcolor[HTML]{FFFFFF}0.0/0.0 & \cellcolor[HTML]{FFFFFF}0.0/0.0 & \cellcolor[HTML]{FFFFFF}0.0/0.0 & \cellcolor[HTML]{FFFFFF}0.0/0.0 \\
  Runr & 100 & Latn, Grek & \cellcolor[HTML]{FFFFFF}0.0/0.0 & \cellcolor[HTML]{FFFFFF}0.0/0.0 & \cellcolor[HTML]{FFFFFF}0.0/1.0 & \cellcolor[HTML]{FFFFFF}0.0/0.0 & \cellcolor[HTML]{FFFFFF}0.0/0.0 & \cellcolor[HTML]{FFFFFF}0.0/0.0 & \cellcolor[HTML]{FFFFFF}0.0/1.0 \\
  Mtei & 100 & Geor, Tibt & \cellcolor[HTML]{FFFFFF}0.0/0.0 & \cellcolor[HTML]{FFFFFF}0.0/0.0 & \cellcolor[HTML]{FFFFFF}0.0/0.0 & \cellcolor[HTML]{FFFFFF}0.0/0.0 & \cellcolor[HTML]{FFFFFF}0.0/0.0 & \cellcolor[HTML]{FFFFFF}0.0/0.0 & \cellcolor[HTML]{FFFFFF}0.0/0.0 \\
  Brai & 100 & Latn, Cyrl & \cellcolor[HTML]{FFFFFF}0.0/0.0 & \cellcolor[HTML]{FFFFFF}0.0/0.0 & \cellcolor[HTML]{FFFFFF}0.0/0.0 & \cellcolor[HTML]{FFFFFF}0.0/0.0 & \cellcolor[HTML]{FFFFFF}0.0/0.0 & \cellcolor[HTML]{FFFFFF}0.0/0.0 & \cellcolor[HTML]{FFFFFF}0.0/0.0 \\
  Cher & 100 & Latn, Grek & \cellcolor[HTML]{FFFFFF}0.0/0.0 & \cellcolor[HTML]{FFFFFF}0.0/0.0 & \cellcolor[HTML]{FFFFFF}0.0/0.0 & \cellcolor[HTML]{FFFFFF}0.0/0.0 & \cellcolor[HTML]{FFFFFF}0.0/0.0 & \cellcolor[HTML]{FFFFFF}0.0/0.0 & \cellcolor[HTML]{FFFFFF}0.0/0.0 \\
  Xpeo & 100 & Latn, Tibt & \cellcolor[HTML]{FFFFFF}0.0/0.0 & \cellcolor[HTML]{FFFFFF}0.0/0.0 & \cellcolor[HTML]{FFFFFF}0.0/0.0 & \cellcolor[HTML]{FFFFFF}0.0/0.0 & \cellcolor[HTML]{FFFFFF}0.0/0.0 & \cellcolor[HTML]{FFFFFF}0.0/0.0 & \cellcolor[HTML]{FFFFFF}0.0/0.0 \\
  Tglg & 100 & Latn, Beng & \cellcolor[HTML]{FFFFFF}0.0/0.0 & \cellcolor[HTML]{FFFFFF}0.0/0.0 & \cellcolor[HTML]{FFFFFF}0.0/0.0 & \cellcolor[HTML]{FFFFFF}0.0/0.0 & \cellcolor[HTML]{FFFFFF}0.0/0.0 & \cellcolor[HTML]{FFFFFF}0.0/0.0 & \cellcolor[HTML]{FFFFFF}0.0/0.0 \\
  Java & 100 & Khmr, Thai & \cellcolor[HTML]{FFFFFF}0.0/0.0 & \cellcolor[HTML]{FFFFFF}0.0/0.0 & \cellcolor[HTML]{FFFFFF}0.0/0.0 & \cellcolor[HTML]{FFFFFF}0.0/0.0 & \cellcolor[HTML]{FFFFFF}0.0/0.0 & \cellcolor[HTML]{FFFFFF}0.0/0.0 & \cellcolor[HTML]{FFFFFF}0.0/0.0 \\
  Bali & 100 & Thai, Khmr & \cellcolor[HTML]{FFFFFF}0.0/0.0 & \cellcolor[HTML]{FFFFFF}0.0/0.0 & \cellcolor[HTML]{FFFFFF}0.0/0.0 & \cellcolor[HTML]{FFFFFF}0.0/0.0 & \cellcolor[HTML]{FFFFFF}0.0/0.0 & \cellcolor[HTML]{FFFFFF}0.0/0.0 & \cellcolor[HTML]{FFFFFF}0.0/0.0 \\
  Phnx & 100 & Latn, Grek & \cellcolor[HTML]{FFFFFF}0.0/0.0 & \cellcolor[HTML]{FFFFFF}0.0/0.0 & \cellcolor[HTML]{FFFFFF}0.0/0.0 & \cellcolor[HTML]{FFFFFF}0.0/0.0 & \cellcolor[HTML]{FFFFFF}0.0/0.0 & \cellcolor[HTML]{FFFFFF}0.0/0.0 & \cellcolor[HTML]{FFFFFF}0.0/0.0 \\
  Linb & 100 & Latn, Hang & \cellcolor[HTML]{FFFFFF}0.0/0.0 & \cellcolor[HTML]{FFFFFF}0.0/0.0 & \cellcolor[HTML]{FFFFFF}0.0/0.0 & \cellcolor[HTML]{FFFFFF}0.0/0.0 & \cellcolor[HTML]{FFFFFF}0.0/0.0 & \cellcolor[HTML]{FFFFFF}0.0/0.0 & \cellcolor[HTML]{FFFFFF}0.0/0.0 \\
  Vaii & 100 & Latn, Mymr & \cellcolor[HTML]{FFFFFF}0.0/0.0 & \cellcolor[HTML]{FFFFFF}0.0/0.0 & \cellcolor[HTML]{FFFFFF}0.0/0.0 & \cellcolor[HTML]{FFFFFF}0.0/0.0 & \cellcolor[HTML]{FFFFFF}0.0/0.0 & \cellcolor[HTML]{FFFFFF}0.0/0.0 & \cellcolor[HTML]{FFFFFF}0.0/0.0 \\
  Tang & 100 & Hani, Latn & \cellcolor[HTML]{FFFFFF}0.0/0.0 & \cellcolor[HTML]{FFFFFF}0.0/0.0 & \cellcolor[HTML]{FFFFFF}0.0/0.0 & \cellcolor[HTML]{FFFFFF}0.0/0.0 & \cellcolor[HTML]{FFFFFF}0.0/0.0 & \cellcolor[HTML]{FFFFFF}0.0/0.0 & \cellcolor[HTML]{FFFFFF}0.0/0.0 \\
  Orkh & 100 & Latn, Hebr & \cellcolor[HTML]{FFFFFF}0.0/0.0 & \cellcolor[HTML]{FFFFFF}0.0/0.0 & \cellcolor[HTML]{FFFFFF}0.0/0.0 & \cellcolor[HTML]{FFFFFF}0.0/0.0 & \cellcolor[HTML]{FFFFFF}0.0/0.0 & \cellcolor[HTML]{FFFFFF}0.0/0.0 & \cellcolor[HTML]{FFFFFF}0.0/0.0 \\
  Brah & 100 & Latn, Hang & \cellcolor[HTML]{FFFFFF}0.0/0.0 & \cellcolor[HTML]{FFFFFF}0.0/0.0 & \cellcolor[HTML]{FFFFFF}0.0/0.0 & \cellcolor[HTML]{FFFFFF}0.0/0.0 & \cellcolor[HTML]{FFFFFF}0.0/0.0 & \cellcolor[HTML]{FFFFFF}0.0/0.0 & \cellcolor[HTML]{FFFFFF}0.0/0.0 \\
  Lisu & 100 & Latn, Cyrl & \cellcolor[HTML]{FFFFFF}0.0/0.0 & \cellcolor[HTML]{FFFFFF}0.0/0.0 & \cellcolor[HTML]{FFFFFF}0.0/0.0 & \cellcolor[HTML]{FFFFFF}0.0/0.0 & \cellcolor[HTML]{FFFFFF}0.0/0.0 & \cellcolor[HTML]{FFFFFF}0.0/0.0 & \cellcolor[HTML]{FFFFFF}0.0/0.0 \\
  Shaw & 100 & Latn, Hebr & \cellcolor[HTML]{FFFFFF}0.0/0.0 & \cellcolor[HTML]{FFFFFF}0.0/0.0 & \cellcolor[HTML]{FFFFFF}0.0/0.0 & \cellcolor[HTML]{FFFFFF}0.0/0.0 & \cellcolor[HTML]{FFFFFF}0.0/0.0 & \cellcolor[HTML]{FFFFFF}0.0/0.0 & \cellcolor[HTML]{FFFFFF}0.0/0.0 \\
  Goth & 100 & Latn, Grek & \cellcolor[HTML]{FFFFFF}0.0/0.0 & \cellcolor[HTML]{FFFFFF}0.0/0.0 & \cellcolor[HTML]{FFFFFF}0.0/0.0 & \cellcolor[HTML]{FFFFFF}0.0/0.0 & \cellcolor[HTML]{FFFFFF}0.0/0.0 & \cellcolor[HTML]{FFFFFF}0.0/0.0 & \cellcolor[HTML]{FFFFFF}0.0/0.0 \\
  Tfng & 100 & Latn, Grek & \cellcolor[HTML]{FFFFFF}0.0/0.0 & \cellcolor[HTML]{FFFFFF}0.0/0.0 & \cellcolor[HTML]{FFFFFF}0.0/0.0 & \cellcolor[HTML]{FFFFFF}0.0/0.0 & \cellcolor[HTML]{FFFFFF}0.0/0.0 & \cellcolor[HTML]{FFFFFF}0.0/0.0 & \cellcolor[HTML]{FFFFFF}0.0/0.0 \\
  Yiii & 100 & Latn, Hani & \cellcolor[HTML]{FFFFFF}0.0/0.0 & \cellcolor[HTML]{FFFFFF}0.0/0.0 & \cellcolor[HTML]{FFFFFF}0.0/0.0 & \cellcolor[HTML]{FFFFFF}0.0/0.0 & \cellcolor[HTML]{FFFFFF}0.0/0.0 & \cellcolor[HTML]{FFFFFF}0.0/0.0 & \cellcolor[HTML]{FFFFFF}0.0/0.0 \\
  Sylo & 100 & Deva, Guru & \cellcolor[HTML]{FFFFFF}0.0/0.0 & \cellcolor[HTML]{FFFFFF}0.0/0.0 & \cellcolor[HTML]{FFFFFF}0.0/0.0 & \cellcolor[HTML]{FFFFFF}0.0/0.0 & \cellcolor[HTML]{FFFFFF}0.0/0.0 & \cellcolor[HTML]{FFFFFF}0.0/0.0 & \cellcolor[HTML]{FFFFFF}0.0/0.0 \\
  Aghb & 100 & Latn, Beng & \cellcolor[HTML]{FFFFFF}0.0/0.0 & \cellcolor[HTML]{FFFFFF}0.0/0.0 & \cellcolor[HTML]{FFFFFF}0.0/0.0 & \cellcolor[HTML]{FFFFFF}0.0/0.0 & \cellcolor[HTML]{FFFFFF}0.0/0.0 & \cellcolor[HTML]{FFFFFF}0.0/0.0 & \cellcolor[HTML]{FFFFFF}0.0/0.0 \\
  Sidd & 75 & Beng, Deva & \cellcolor[HTML]{FFFFFF}0.0/0.0 & \cellcolor[HTML]{FFFFFF}0.0/0.0 & \cellcolor[HTML]{FFFFFF}0.0/0.0 & \cellcolor[HTML]{FFFFFF}0.0/0.0 & \cellcolor[HTML]{FFFFFF}0.0/0.0 & \cellcolor[HTML]{FFFFFF}0.0/0.0 & \cellcolor[HTML]{FFFFFF}0.0/0.0 \\
  
\bottomrule
\end{tabular}
}%resizebox
\end{table}

% \newpage
% \clearpage
% \input{checklist.tex}

\end{document}